\documentclass[12pt]{article}
\usepackage{sdss2020} % Uses Times Roman font (either newtx or times package)
\usepackage[british]{babel}
\usepackage{url}
\usepackage{latexsym}
\usepackage{amsmath, amsthm, amsfonts}
\usepackage{mathtools}

\DeclarePairedDelimiter\floor{\lfloor}{\rfloor}
\usepackage{algorithm, algorithmic}  
\usepackage{subfig}
\usepackage{graphicx}
\usepackage{caption}
\usepackage{natbib}
\usepackage{pdfpages}
\usepackage{grffile}
\usepackage{newtxtext}
\usepackage{newtxmath}
\usepackage{hhline}
\usepackage{anyfontsize}
\usepackage[T1]{fontenc}
\usepackage{textcomp}

\usepackage{hyperref}

\title{AR-Sieve Bootstrap for the Random Forest and a simulation-based comparison with rangerts time series prediction}

 \author{%
Cabrel Teguemne Fokam\textsuperscript{$\dagger$}, Carsten Jentsch\textsuperscript{$\dagger$}, 
Michel Lang\textsuperscript{*}, Markus Pauly\textsuperscript{$\dagger$,*}\\[1ex]
$\dagger$ Department of Statistics, TU Dortmund University, Germany \\
* Research Center Trustworthy Data Science and Security, UA Ruhr, Germany
}

\begin{document}
\bibliographystyle{unsrtnat} % unsrt Please do not change the bibliography style
%Where the bibliography will be printed
%\null

%\includepdf[]{pdfs/cover_page.pdf}
%\input{title_page}
%\includepdf{pdfs/Eidesstattliche_Versicherung_signed.pdf}
%\input{general_info.tex}

\maketitle

\begin{abstract}
The Random Forest (RF) algorithm can be applied to a broad spectrum of problems, including time series prediction. However, neither the classical IID (Independent and Identically distributed) bootstrap nor block bootstrapping strategies (as implemented in rangerts) completely account for the nature of the Data Generating Process (DGP) while resampling the observations. We propose the combination of RF with a residual bootstrapping technique where we replace the IID bootstrap with the AR-Sieve Bootstrap (ARSB), which assumes the DGP to be an autoregressive process. To assess the new model's predictive performance, we conduct a simulation study using synthetic data generated from different types of DGPs. It turns out that ARSB provides more variation amongst the trees in the forest. Moreover, RF with ARSB shows greater accuracy compared to RF with other bootstrap strategies. However, these improvements are achieved at some efficiency costs.
\end{abstract}


\section{Introduction}
%one of the most relied on models ..
Random Forest (RF) \citep{rf} has become one of the go-to models for data analysis because of its flexibility and performance. RF also appears to perform well in time series forecasting \citep{huang2020travel, kane2014comparison, naing2015forecasting}.
%In general, the accuracy of a RF is related to the correlation amongst its trees \citep{hastie2009, Lee2020}

The relationship between the accuracy of RF and the correlation among its trees has been established \citep{hastie2009, Lee2020}. The less correlated the trees, the more accurate the model. In a seminal paper, \cite{rf} used classical bootstrapping \citep{efron1979bootstrap}, also known as IID Bootstrap, to construct the trees.

However, when dealing with time series data, the IID assumption is not valid since the observations are dependent by nature, and thus their dependency could be broken. \cite{rf_block_bootstrap} proposed replacing the IID bootstrap with different block bootstrap strategies \citep{nbb, mbb, sbb, cbb} and coined the new method rangerts. % They showed that their approach can be better that the original RF.
They exemplified that this approach can be better than the standard RF. However, they only used two benchmark datasets and an extensive simulation study confirming these findings is still missing. Moreover, although IID and block bootstrapping work well in practice, they do not consider the complete nature of the DGP.

In this paper, we propose to use the AR-Sieve Bootstrap (ARSB) \citep{ar_sieve, kreiss1988asymptotic} instead of the IID bootstrap to construct the trees of the RF. The ARSB draws the bootstrap samples from a fitted autoregressive (AR) model and has already been shown to perform well for other time series models such as ARMA (AutoRegressive Moving Average) models \citep{kreiss2011range}. 
To assess the performance of this new RF model, we compare its predictive accuracy with that of five RFs variants and a benchmark model for time series forecasting based on an autoregressive model fit in extensive simulations.
%Thereby we consider 
For this purpose, we consider six different classes for the DGP: AR- \citep{arima}, MA- (Moving Average, \citep{arima}), ARMA- \citep{arima}, ARIMA- (AutoRegressive Integrated Moving Average, \citep{arima}), ARFIMA- (AutoRegressive Fractionally Integrated Moving Average, \citep{arfima}), and GARCH (Generalized AutoRegressive Conditional Heteroskedastic, \citep{garch}) processes, see sections \ref{bootstrap_strategies} and \ref{SIMULATION} for the explicit definition of the DGPs.

We start with a brief introduction to RF in Section \ref{RF} and its different bootstrap strategies used in the literature. We then present the new approach with the ARSB and compare its %time and space 
computational complexity with that of the other bootstrap methods in Section \ref{bootstrap_strategies}. Section \ref{SIMULATION} concludes with the results and an outlook.% of a simulation study.
%The rest of this present an overview of the different bootstrap strategies, section  . In the last section a simulation study

%To achieve that goal, Breiman \citep{rf} used in the original paper on Random Forest IID Bootstrapping \citep{iid} of the original dataset to construct the forest trees. 

%The analysis of time series data . Random Forest \citep{rf} has already been 

%Time Series is a well known and studied field which appear in a wide range of application: Economics, Finance, Industry, etc. The emergence of machine learning brought new models to 

%The analysis of time dependent gave birth to several parametric models.

%The evolution of the Machine Learning brought new ideas and new models have been developed, Random Forest (RF) \citep{rf} among them. 

\section{Random Forest}
\label{RF}

The Random Forest learning algorithm is a bagging (Bootstrap  aggregating) \citep{breiman1996bagging} technique. A bagging algorithm merges the predictions of multiple base learners to obtain a better prediction than its individuals. 
The more diverse the learners, the more accurate the ensemble. As its name suggests, RF uses decision trees \citep{breiman2017tree} as base learners. As a first step, variability is achieved in RF  through bootstrapping, which determines the observations to be fed for each tree construction.

A tree is built by recursively splitting the observations of a node into two disjoint partitions, starting from the root node, which contains all the bootstrap observations. Only a randomly chosen subset of size \textit{mtry} of the features is considered for the split. This is the second source of diversity in RF. 
It is either tuned as a hyperparameter during training or chosen from the default settings: For a dataset with $p$ features, the default choice for \textit{mtry} is $\sqrt{p}$ in classification  and $p/3$ for regression tasks. The best split is done at the point along one of the \textit{mtry} features' axes which minimises the average impurity of the resulting child nodes. In a regression context, the impurity is quantified via the variance \citep{breiman2017tree} -- the lower the variance, the purer the node. Other impurity criteria, such as the least absolute deviation \citep{rf_least_absolute_deviation}, can also be used. 
The tree construction has to be stopped once some criteria are met to avoid over-fitting. Some of those criteria include the \textit{minimum node size} to attempt a split or the \textit{maximum depth} of the tree. Once built, the final prediction is obtained by averaging the individual predictions of each tree. The \texttt{ranger} package \citep{ranger} provides a fast implementation of RF with a wide choice of parameters to tune.

\section{Bootstrap strategies for Random Forest}
\label{bootstrap_strategies}

%The different block bootstrap strategies used in the literature are:
%\begin{itemize}

  %  \item IID Bootstrap \citep{efron1977bootstrap}, and
   % \item(Moving) Block Bootstrap (MBB) \citep{mbb}.
   % \item Stationary Block Bootstrap (SBB)  \citep{sbb},
   % \item Circular Block Bootstrap (CBB) \citep{cbb}, 
    
%\end{itemize}
%We explain these methods together with the ARSB subsequently.

There exist different bootstrap strategies for the RF in the literature. The most common one is the IID bootstrap \citep{efron1979bootstrap} which is implemented in the \texttt{ranger} package. Moreover, for time series forecastsing several block bootstrap strategies have been suggested recently \citep{rf_block_bootstrap} and were implemented in the \texttt{rangerts} package \citep{rangerts}. We explain the latter together with the ARSB in the sequel. To this end, we consider a time series model of length $T$ given by real-valued random variables $Y_t$, $t=1,\ldots, T$.

%AR-Sieve Bootstrap strategy makes an assumption about the nature of the DGP, which might not be correct in most cases but capture some of its properties.

\subsection{Block bootstrapping}

%Non-parametric bootstrapping strategies for RF applied to time series data use blocks. They try to keep some portion of the data together not to break their dependency.
The main idea of block bootstrapping is to keep some portion of the data together in the form of blocks to avoid breaking their dependency \citep{mbb}.

\begin{figure}[H]
    \centering
\captionsetup{justification=centering}
    
    \includegraphics[width=0.5\textwidth]{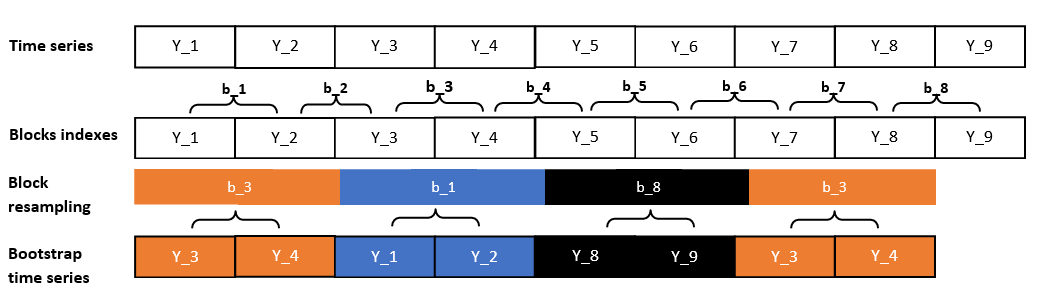}
\caption{\textit{Moving Block Bootstrap for a time series of length $T = 9$ with $\ell = 2$ as block length. Here, we have $B = 8$ blocks from which $k=4$ blocks are drawn with replacement.}}%
    \label{fig:mbb}
\end{figure}

\cite{kunsch1989jackknife}
proposed the so-called Moving Block Bootstrap (MBB). Here, the time series is first divided into $B$ overlapping blocks of length $\ell$, $B= T-\ell+1$, where the block 
%das ist so leider nicht korrekt mit den gewählten Indizes $b_i = (Y_{(i-1)\ell + 1}, \ldots, Y_{i\ell})$ 
$b_i = (Y_{i}, \ldots, Y_{i+\ell-1})$ 
starts at time index $i$, $i = 1,\ldots, B$. Then $k = \floor*{\frac{T}{\ell}}$ blocks are drawn independently with replacement and joined together in the order in which they were drawn to recover the original length of the time series. Figure \ref{fig:mbb} illustrates this strategy for a time series of length $T = 9$ using a block length of $\ell = 2$.

%Moving block bootstrap divides the series in $k = \frac{T}{l}$ chunks of length $l$ each. $k$ chunks are then sampled with replacement to form a new time series of length $T$.\\

%MBB has some derivatives where the blocks' length are different (Stationary Block Bootstrap \citep{sbb}), where the time series has a circular shape (Circular Block Bootstrap \citep{cbb}) or simply where the blocks have no observations in common (Non-Overlapping Block Bootstrap \citep{nbb}).

The MBB is easy to implement but also has some drawbacks. Beyond the choice of a 'good' block length, it neglects the dependency between blocks in the and the bootstrap sample will in general not be stationary, see e.g. \cite{politis1994stationary}. Several variants exist that partially solve these problems: The Stationary Block Bootstrap (SBB, \citep{sbb}) allows for different block lengths, while the Circular Block Bootstrap (CBB, \citep{cbb}) assumes the time series to have a circular shape, and the Non-Overlapping Block Bootstrap (NBB, \citep{nbb}) builds blocks with no common observations.

%The blocks can have observations in common (Moving Block Bootstrap) or not (Non-Overlapping Block Bootstrap \citep). 

%In NBB, if $T$ is not a multiple of $\ell$, that is $T \equiv r\mod \ell \text{,\hspace{2mm}$r\neq 0$} $ , the first $r$ observations of the series can be removed before performing the Bootstrap.

%Stationary Block Bootstrap can be seen as a special case of Moving Block Bootstrap where the the blocks length are variable. The block lengths $\ell_i$, $i=1,\dots,k$ are randomly sampled from the geometric distribution \citep{block_length_selection} and $\sum_{i=1}^k \ell_i = T$

%If the blocks are of variable lengths $l_1, l_2, ..., l_k $ we obtain stationary bootstrap. \\

%Circular Block Bootstrap can be also considered as a special case of Moving Block Bootstrap. Here the series has a circular shape, that is, the last observations $Y_T$ is followed by the first observation $Y_1$. There is no real beginning nor end of the time series.

Nevertheless, the IID Bootstrap is the benchmark resampling strategy used for RF and can be seen as a specialisation of the above block bootstrap strategies in which the blocks' lengths are chosen as 
$\ell = 1$. %$\ell_i = 1$.

\subsection{AR-Sieve Bootstrap}

The AR-Sieve bootstrap (ARSB) \citep{ar_sieve, kreiss1988asymptotic, kreiss2011range} uses residual resampling by fitting an autoregressive process on the data. The fitted model is linear and additive in the noise term with the following form for $t = p+1, \ldots, T$:

\begin{equation}
    Y_t-\mu = \sum_{j=1}^{p}\phi_j (Y_{t-j}-\mu) + \epsilon_t ,
\end{equation}

%$$ Y_t-\mu = \sum_{j=1}^{p}\phi_1 (Y_{t-j}-\mu) + \epsilon_t, \textit{t = p+1 ,..., $T$} $$

where $\mu = E(Y_t)$ is the mean of the stationary time series, $p$ the order of the model, $\phi_i$ with $i = 1, \ldots ,p$ the model coefficients, and $\epsilon_t \sim N(0,\sigma^2)$ the white noise process. The ARSB consists of four steps:
\begin{enumerate}
    \item Fit the model and obtain the estimated Yule-Walker coefficients $\hat{\phi}_1,...,\hat{\phi}_p$ and residuals $\hat{\epsilon}_t$,
    \item Center the residuals around 0 if the fitted model has no intercept: $\hat{\epsilon}_t^{'} = \hat{\epsilon}_t - \Bar{\hat{\epsilon}}_t$,
    \item Draw from (centred) residuals $\epsilon_t^*$ with replacement,
    \item Construct $Y_t^* = \hat{\mu} + \sum_{j=1}^p \hat{\phi}_j (Y_{t-j}^*-\hat{\mu}) + \epsilon_t^*$.
\end{enumerate}

In this paper, we use the Levinson-Durbin recursion \citep{levinson_durbin} to fit the model because it ensures the resulting bootstrap series $Y^*$ to be stationary \citep{kreiss2011range}.  % Ordinary Least Squares (OLS), Maximum Likelihood Estimation (MLE) or similar methods can also be used.
To determine the order of the fitted model, we use Akaike's Information Criterion (AIC), which is asymptotically equivalent to the Leave-One-Out Cross-Validation for the model's selection \citep{shao1993}. Moreover, AIC is not time-consuming. It is particularly suited for simulations as no manual checking is required. In practice, other methods or criteria, e.g. based upon Auto-Correlation Functions (ACFs) plots, can be used.

Although the fitted model is linear with Gaussian errors, the approach is theoretically valid for more general DGPs, see % stationary processes that posses a so-called general Wold-type autoregressive representation, see
\cite{kreiss2011range} for details. We evaluate the use of ARSB in the RF for forecasting different type of linear and non-linear processes in the simulation study in Section~\ref{SIMULATION}. Before that we shortly discuss the computational complexity of all bootstrap methods.

\subsection{Computational complexity}

\begin{figure}[H]
    %\centering
    %\captionsetup{justification=centering}
    
    \includegraphics[width=0.5\textwidth, height =4cm]{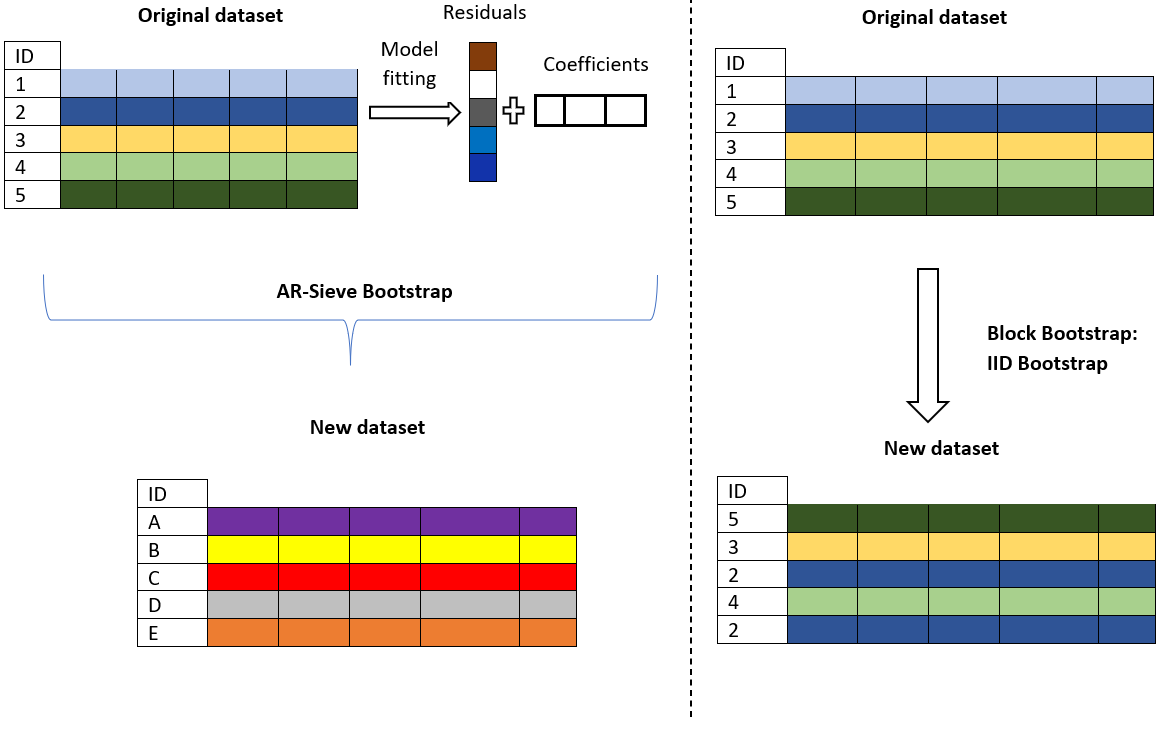}
\caption{\small \textit{\textbf{left}: ARSB generates the new dataset from the original one. The new dataset usually has no common observation with the original dataset.\newline
                 \textbf{right}: the new dataset is created using IID Bootstrap: the first observation has not been sampled. Only the indices [5,3,2,4,2] need to be saved, the bootstrapped samples are also found in the original dataset.}}%
    \label{fig:complexity}
\end{figure}

Block bootstrap methods are all index-based. Only the samples' indices need to be known for a bootstrap dataset to be created, making block bootstrapping strategies efficient. They typically perform in $\mathcal{O}(T)$ time and only need $\mathcal{O}(T)$ space (Figure \ref{fig:complexity}, right) for each tree to be built.
%$\mathcal{O}(n\log{}n)$

On the contrary, ARSB is less efficient since it is a residual resampling technique. 
The model fitting can take up to $\mathcal{O}(p^2)$ \citep{levinson_durbin} operations to solve the Yule-Walker (YW) equations using the Levinson-Durbin recursion, and $\mathcal{O}(T*p)$ time to reconstruct the time series. In practice, $p \ll T$, such that ARSB can be executed in linear time $\mathcal{O}(T*p)$.

The AR-Sieve strategy also requires more memory space. The bootstrap time series and its first $p$ lags need to be stored (Figure \ref{fig:complexity}, left), requiring $\mathcal{O}(T*p)$ space. However, the additional effort required by the ARSB may prove beneficial as it helps to create more diverse trees, which may increase the RF's accuracy. Whether this intuition is really true will be evaluated in the following section.

\section{Simulation study}
\label{SIMULATION}
\subsection{Setup}

To compare the models' performances, we conduct a Monte Carlo experiment with $M = 1000$ iterations. Data are generated from AR-, MA-, ARMA-, ARIMA- and ARFIMA as well as GARCH processes with Gaussian white noise  $\epsilon_t \sim N(0,1)$. The first five DGPs are described in the most general (ARFIMA) form as
\begin{equation}
    (Y_t - \sum_{i=1}^{p}\phi_i Y_{t-i})(\sum_{k=1}^{\infty} \binom{d}{k} (-1)^k Y_{t-k})= \epsilon_t - \sum_{j=1}^{q}\theta_j \epsilon_{t-j}
    \label{eq:general_arima}
\end{equation}
while GARCH processes are generated via
$$
    \begin{aligned}
        Y_t &= \sigma_t \epsilon_{t},\\
        \sigma_t &= \alpha_0 + \sum_{i=0}^{p}\alpha_i\epsilon_{t-i}^2 + \sum_{i=j}^{q}\beta_j\sigma_{t-j}^2 %\\
%        \epsilon_t &\sim N(0,1) 
    \end{aligned}
\label{eq:general_arch}
$$

For each DGP we consider different parameter configurations as given below:
\begin{itemize}
    \item AR(1): $\phi_1 \in $\{0.2, -0.2, 0.5, -0.5, 0.8, -0.8\},
    \item MA(1): $\theta_1 \in $\{0.2, -0.2, 0.5, -0.5, 0.8, -0.8\},
     \item ARMA(1,1):$\{(\phi_1 \!=-0.4,\theta_1=-0.2), (\phi_1 =-0.3,\theta_1=0.4), (\phi_1=0.1,\theta_1=0.3),(\phi_1=0.1,\theta_1=0.7), (\phi_1=0.7,\theta_1=0.1), (\phi_1=0.7,\theta_1=0.1) \}$,
    
    \item ARIMA(1,1,1):$\{(\phi_1=0.1,\theta_1=0.3), (\phi_1=0.7,\theta_1=0.1), (\phi_1=0.1,\theta_1=0.7) \}$,
    
    \item ARFIMA(1,0.3,1): $\{(\phi_1 = 0.3, \theta_1 = 0.4, d = 0.3), (\phi_1 = 0.7, \theta_1  = 0.2, d = 0.3)\}$,
    %$\{\phi_1 = 0.7, \theta_1  = 0.2, d = 0.3\}$
    
    \item GARCH(1,1): $\{(\alpha_0 = 0.01, \alpha_1 = 0.3, \beta_1 = 0.6),(\alpha_0 = 0.01, \alpha_1 = 0.05, \beta_1 = 0.9)\}$.
   
\end{itemize}
We thus consider twenty-five (25) %four (24) 
time series DGPs, each in three different sizes $T \in$ \{{100, 500 and 1000}\}, yielding a total of $75$ %72 
parameters configurations. 
%where where $B$ is the Backshift operator: $$B^d Y_t = Y_{t-d}$$
%$d \in \mathbb{R}$
 For each configuration, we find the Yule-Walker estimates for the ARSB coefficients and train RFs with the bootstrap strategies presented in Section~\ref{bootstrap_strategies}. One-step ($h=1$) and five-step ahead ($h=5$) predictions are then made using the recursive multi-step forecast \citep{taieb2012recursive} method. At each iteration, data is generated, the different models are fitted, and their performances are evaluated via the Mean Square Error (MSE). To obtain a unified metric over the $M$ iterations, % to assess the experiment,
 we use the median of all MSEs %. Other metrics, such as the mean of MSEs, can also be used 
 \citep{metrics}. 

The simulations are realised with the \texttt{R} programming language \citep{R}, and the existing RF models are created using the \texttt{rangerts} package \citep{rangerts}, which is an extension of the \texttt{ranger} package \citep{ranger} for time series prediction. We perform no tuning and use the parameters' default values provided by the package. We further extended \texttt{rangerts} to support ARSB and made the extension accessible on \href{https://github.com/tfcyoo/rangerts_ars}{GitHub}.

\subsection{General results}

Only a summary of the results is presented to avoid redundancy. Detailed results can be found in appendices \ref{app_h_1} and \ref{app_h_5}.% for one-step and five-step ahead predictions.

\begin{figure}[H]
    \centering
    \captionsetup{justification=centering}
    
    \includegraphics[width=0.5\textwidth, ]{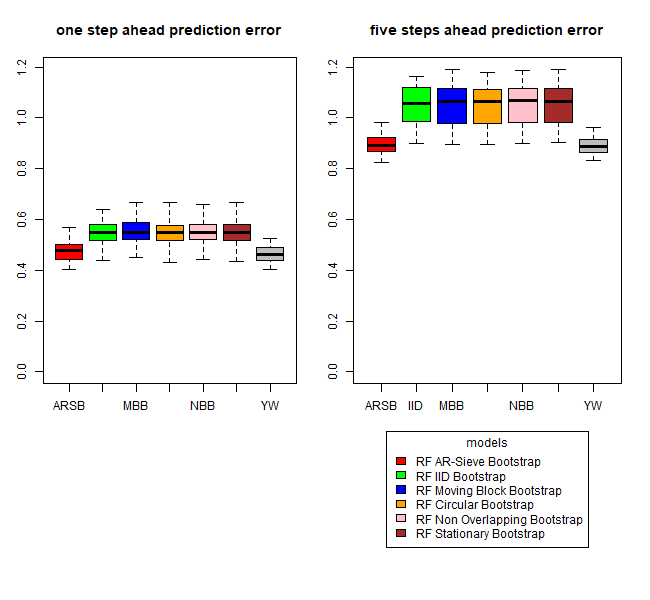}
    
\caption{\textit{\small Box plots of the Median of MSEs the models across the simulations for the six classes of DGP and for $h=1$ (\textbf{left}) and $h=5$ (\textbf{right})}.}%
    \label{fig:horizon_compare}
\end{figure}

%a bell-shaped curve 
We observe an improvement for the new RF of up to 13\% and 16\% for one-step and five-step ahead prediction, respectively (Figure \ref{fig:horizon_compare}) for the median of MSEs  compared to the other RF variants. ARSB successfully creates more diverse trees than its counterparts. The new RF's performance is comparable with that of the YW estimator. This similarity shows that the properties of the fitted AR model have been conserved during tree construction with ARSB. However, it performs less good if the DGP has a high value on the MA part of Equation \ref{eq:general_arima} (Figures \ref{fig:ma_5_h_1}-\ref{fig:ma_6_h_1}, \ref{fig:arima_1_h_1}-\ref{fig:arima_3_h_1}) but still performs better than the other RF models.
%On the box-plots of figure \ref{fig:horizon_compare} 

%To obtain a models' performances summary for the simulations, we draw box-plots (Figure \ref{fig:horizon_compare}) of the median of MSEs across all the simulations. The new RF and YW estimator show similar performance for both short and long term prediction in terms of accuracy and 

%We observe the plots exhibit a square wave form for both $h = 1$ and $h = 5$, implying that the combination RF ARSB and Yule-Walker (YW) estimator perform best in terms of accuracy and variance. Interestingly, the variance of the error remains quite the same for the new RF in one step and five steps ahead predictions, showing good stability as opposed to the other RF models.  

%The results' similarities between those two models show that most of the properties of the fitted AR process were conserved by the ARSB in RF. ARSB therefore successfully helps create more diverse, less correlated trees than the other bootstrap strategies. 

%However, RF with ARSB appears to struggle when the DGP has a high parameter value on the MA part % and with ARIMA processes in general 
%for both short and long term prediction (figures \ref{fig:ma_5_h_1}-\ref{fig:ma_6_h_1}, \ref{fig:arima_1_h_1}-\ref{fig:arima_3_h_1}). But still it does better than the other RFs.

The models' performances were also ranked (tables \ref{table:ranking_h_1} and \ref{table:ranking_h_5}) on each of the 72 simulation configurations and DGP. Ties were resolved using mean ranks.

\begin{table}[H]

    \small

    \begin{tabular}{|c c c c c c c c|} 
    
         \hline
          DGP & ARSB & IID &  MBB & CBB & NBB & SBB & YW\\ 
         \hline
         \scriptsize AR& 1.61 &5.06 &4.67 &4.72 &5.67 &4.89 &\textbf{1.39} \\
         %\hline
           \scriptsize MA&1.72 &4.61 &5.17 &4.72 &5.11 &5.22 &\textbf{1.44} \\
        %\hline

       \scriptsize ARMA& 1.6 &4.53 &5.47 &5.33 &4.6 &5.07 &\textbf{1.4}\\
        %\hline
   \scriptsize GARCH&\textbf{ 1.33 }&4.33 &5.5 &3.67 &5.83 &5.67 &1.67 \\
    %\hline
\scriptsize ARFIMA&1.83 &4.5 &5.33 &4.83 &5.5 &4.83 &\textbf{1.17}\\
%\hline
\scriptsize ARIMA&1.67 &5 &5.33 &4.33 &4.67 &5.67 &\textbf{1.33} \\
\hline
\scriptsize Overall & 1.64 &4.72& 5.17& 4.72 &5.18 &5.17 &\textbf{1.40} \\
\hline
        
    \end{tabular}
    \caption{Models' Average ranking for $h = 1$.}
    \label{table:ranking_h_1}
\end{table}

\begin{table}[H]

    \small

    \begin{tabular}{|c c c c c c c c |} 
    
         \hline
          DGP & ARSB & IID &  MBB & CBB & NBB & SBB & YW\\ 
         \hline
         \scriptsize AR&\textbf{1.36} &4.5 &5.17 &4.5 &5.28 &5.56 &1.64\\
         %\hline
           \scriptsize MA&1.61 &4.39 &5.39 &4.56 &5.17 &5.5 &\textbf{1.39}\\
        %\hline

       \scriptsize ARMA&\textbf{1.47}&4.47 &5.4 &4.33 &5.13 &5.67 &1.53\\
        %\hline
   \scriptsize GARCH&\textbf{1.5} &4.17 &5.17 &4.33 &5.33 &6 &\textbf{1.5} \\
    %\hline
\scriptsize ARFIMA&1.83 &4.67 &5.67 &3.5 &5.83 &5.33 &\textbf{1.17}\\
%\hline
\scriptsize ARIMA&1.78 &4.33 &4.56 &5.22 &5.44 &5.44 &\textbf{1.22}\\
\hline

\scriptsize Overall & 1.55& 4.43& 5.24& 4.47& 5.29 &5.57 &\textbf{1.45} \\

 \hline
    \end{tabular}
    \caption{Models' Average ranking for $h = 5$.}
    \label{table:ranking_h_5}
\end{table}

Although the new RF was not the best performing model overall, the results are still encouraging given that no parameter tuning was done. Moreover, different to the classical YW time series approach, RF can also support exogenous variables. It also performs better on long-term (h=5) than short-term prediction (h=1) and when the AR part dominates the DGP (figures \ref{fig:ar_5_h_1} - \ref{fig:ar_6_h_1}).

Overall, Benchmark RF and RF with block bootstrap show similar performances on non-seasonal time series, contrary to the Goehry et al. \citep{rf_block_bootstrap} study, where the time series were seasonal. This hints at better adequacy of block bootstrapping for seasonal time series.

\begin{figure}[H]
    \centering
    \captionsetup{justification=centering}
    
    \includegraphics[width=0.5\textwidth]{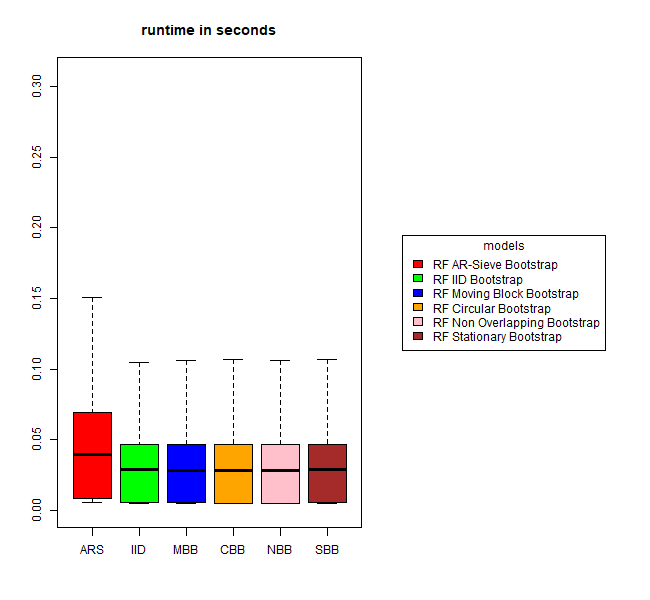}
\caption{\textit{\small Box-plots of models' average runtime across the simulations for the six classes of DGP.}}%
    \label{fig:time_compare}
\end{figure}

Regarding the running time, RF with ARSB has the highest value (Figure \ref{fig:time_compare}) as expected and was approximately up to two times slower than the other RF models. Nevertheless, its running time does not increase linearly with the order of the fitted model (table \ref{table:runtime}). It remains low enough with an average value of $0.06$ seconds for an average time series length of $T = 533$ and an average fitted order of $p = 3.4$.

\begin{table}[H]

    \small

    \begin{tabular}{|p{8mm} c c c c c c c|}
    
         \hline
          DGP & P &ARSB & IID &  MBB & CBB & NBB & SBB \\ 
        \hline
         \scriptsize AR&1.81 &0.03 &0.02 &0.02 &0.02 &0.02 &0.02 \\
        % \hline
           \scriptsize MA& 4.70&0.06 &0.04 &0.04 &0.04 &0.04 &0.04 \\
       % \hline

       \scriptsize ARMA& 3.21&0.04 &0.03 &0.03 &0.03 &0.03 &0.03 \\
       % \hline
   \scriptsize GARCH& 2.76&0.03 &0.03 &0.03 &0.03 &0.03 &0.03 \\
   % \hline
\scriptsize ARFIMA&3.76 &0.06 &0.04 &0.04 &0.04 &0.04 &0.04 \\
%\hline
\scriptsize ARIMA&4.45 &0.12 &0.07 &0.07 &0.07 &0.07 &0.07 \\
\hline

\scriptsize Overall&3.4 & 0.06 &0.04 &0.03& 0.03& 0.04& 0.04 \\

 \hline
    \end{tabular}
    \caption{Models' Average runtime per model.}
    \label{table:runtime}
\end{table}

\section{Conclusion and Perspectives}

We introduced the AR-Sieve bootstrap (ARSB) method for the random forest (RF) algorithm, a residual resampling strategy based upon a fitted AR process. A simulation study on synthetic data showed better forecasting accuracy of the proposed approach compared to IID and Block Bootstrap strategies. However, ARSB appeared to be computationally more demanding than its counterparts but remained fast enough for practical applications. Moreover, RF with ARSB appears to conserve most properties of an AR process but struggles with DGPs having high coefficients on the MA part. 

The conducted study was empirical and a more detailed and theoretical study regarding validity and consistency to support these findings needs to be done. This could also help to find out whether extensions of the ARSB \citep{fragkeskou2018extending} are worthwhile for the RF. A similar study with additional hyperparaemter tuning \citep{bischl2017mlrmbo} as well as exogenous information, e.g., 
on several benchmark datasets, would also provide a better insight into the model's performance. 

%Another exciting development would be considering ARSB for high-dimensional time series data using factor modelling proposed by Bi et al. \citep{bi2021ar}. This method may increase efficiency while conserving desired properties like the stationarity of the bootstrap time series. 

\bibliography{references}

\appendix

\section{One step ahead forecast}
\label{app_h_1}
\subsection{AR Models}

\begin{figure}[H]
    \centering
    \captionsetup{justification=centering}
    
    \includegraphics[width=0.5\textwidth, height =6cm]{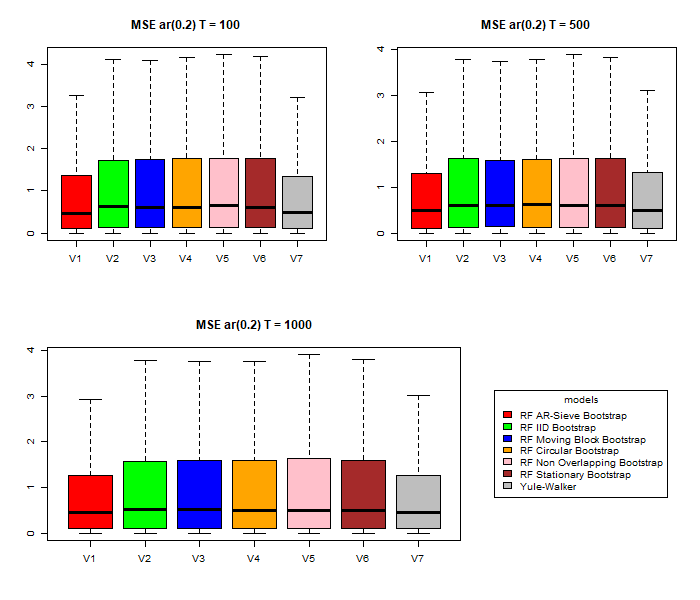}
\caption{\textit{\small AR(1): $\phi_1 = 0.2$}}%
    \label{fig:ar_1_h_1}
\end{figure}

\begin{figure}[H]
    \centering
    \captionsetup{justification=centering}
    
    \includegraphics[width=0.5\textwidth, height = 6cm]{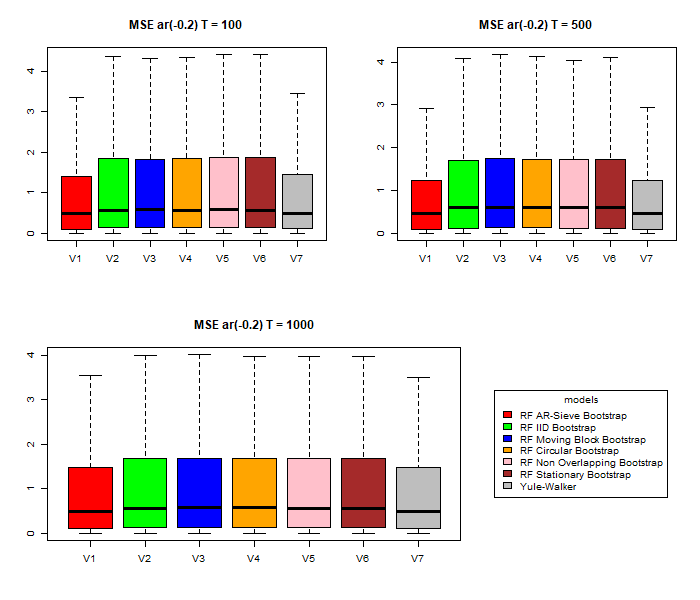}
\caption{\textit{\small AR(1): $\phi_1 = -0.2$}}%
    \label{fig:ar_2_h_1}
\end{figure}

\begin{figure}[H]
    \centering
    \captionsetup{justification=centering}
    
    \includegraphics[width=0.5\textwidth, height=6cm]{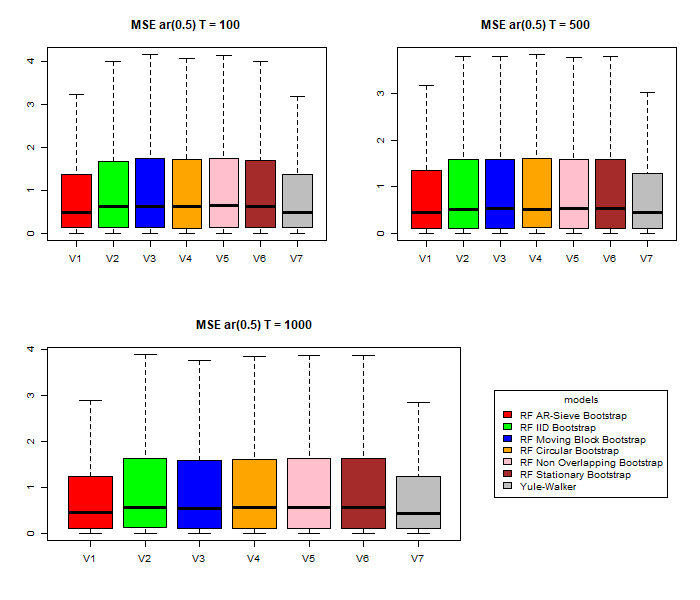}
\caption{\textit{\small AR(1): $\phi_1 = 0.5$}}%
    \label{fig:ar_3_h_1}
\end{figure}

\begin{figure}[H]
    \centering
    \captionsetup{justification=centering}
    
    \includegraphics[width=0.5\textwidth, height=6cm]{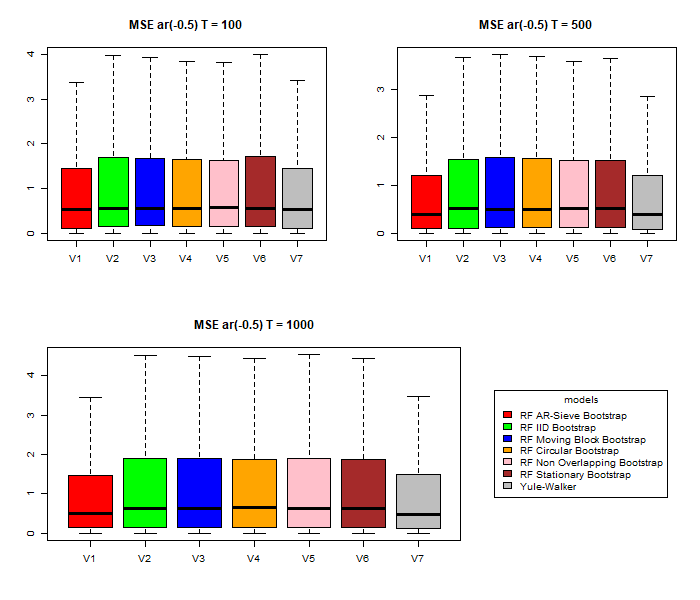}
\caption{\textit{\small AR(1): $\phi_1 = -0.5$}}%
    \label{fig:ar_4_h_1}
\end{figure}

%-------------AR high coefficients

\begin{figure}[H]
    \centering
    \captionsetup{justification=centering}
    
    \includegraphics[width=0.5\textwidth,height=6cm]{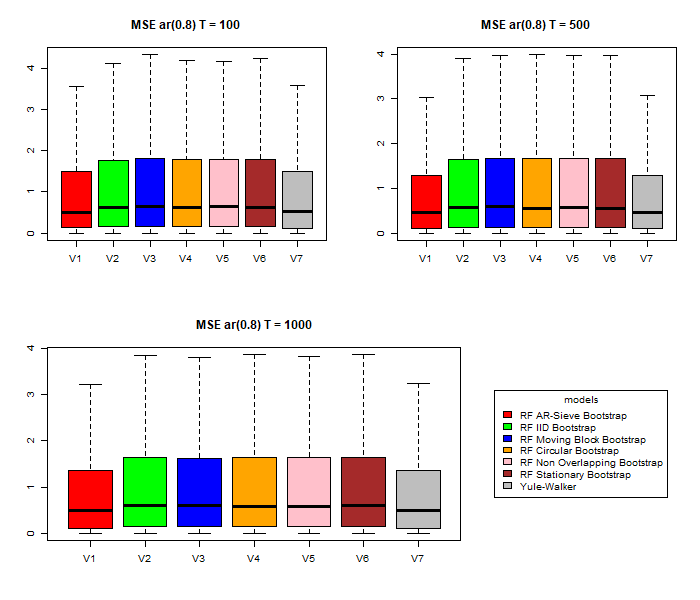}
\caption{\textit{\small AR(1): $\phi_1 = 0.8$}}%
    \label{fig:ar_5_h_1}
\end{figure}

\begin{figure}[H]
    \centering
    \captionsetup{justification=centering}
    
    \includegraphics[width=0.5\textwidth,height=6cm]{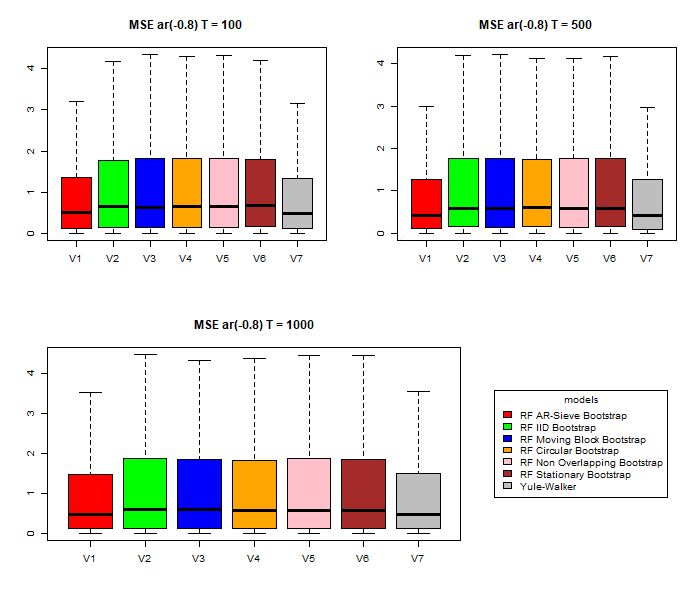}
\caption{\textit{\small AR(1): $\phi_1 = -0.8$}}%
    \label{fig:ar_6_h_1}
\end{figure}

%%%%%%%%%%%%%%%%%%%%%%%%%%%%%%% MA

\subsection{MA Models}
%------------ MA small coefficients

\begin{figure}[H]
    \centering
    \captionsetup{justification=centering}
    
    \includegraphics[width=0.5\textwidth,height=6cm]{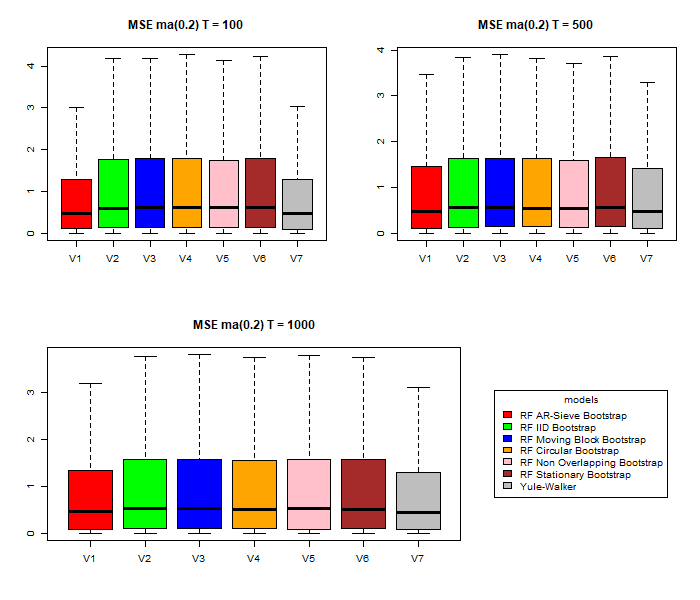}
\caption{\textit{\small MA(1): $\theta_1 = 0.2$}}%
    \label{fig:ma_1_h_1}
\end{figure}

\begin{figure}[H]
    \centering
    \captionsetup{justification=centering}
    
    \includegraphics[width=0.5\textwidth,height=6cm]{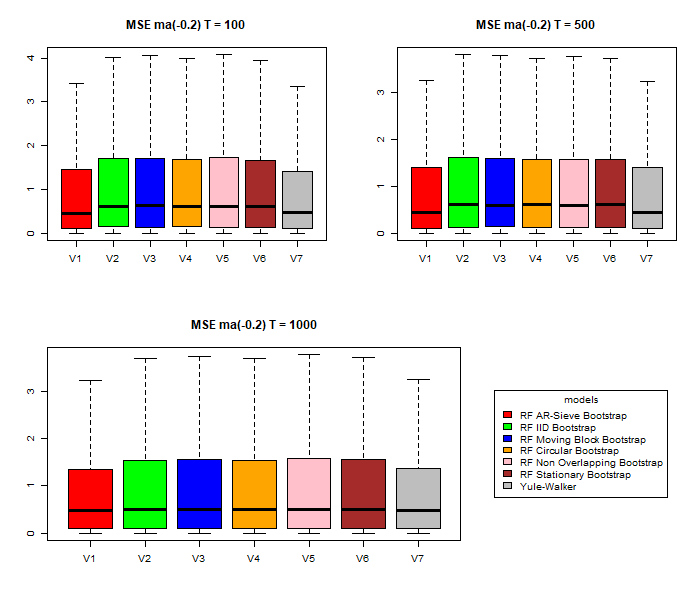}
\caption{\textit{\small MA(1): $\theta_1 = -0.2$}}%
    \label{fig:ma_2_h_1}
\end{figure}

\begin{figure}[H]
    \centering
    \captionsetup{justification=centering}
    
    \includegraphics[width=0.5\textwidth,height=6cm]{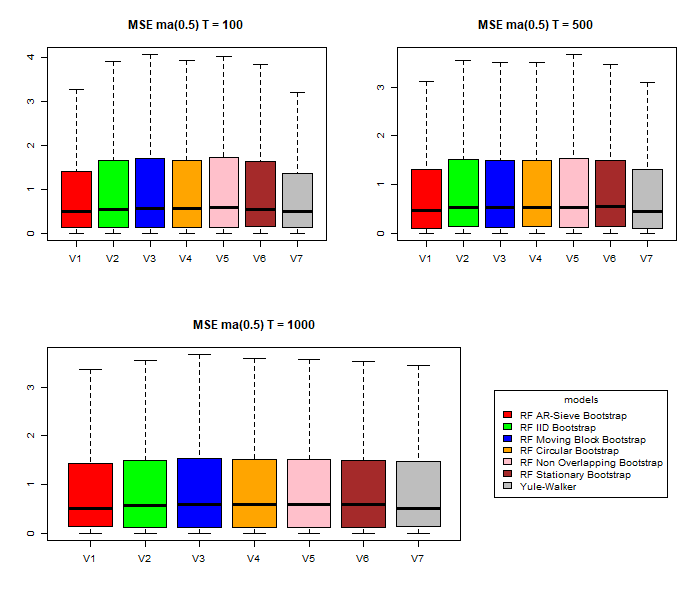}
\caption{\textit{\small MA(1): $\theta_1 = 0.5$}}%
    \label{fig:ma_3_h_1}
\end{figure}

\begin{figure}[H]
    \centering
    \captionsetup{justification=centering}
    
    \includegraphics[width=0.5\textwidth,height=6cm]{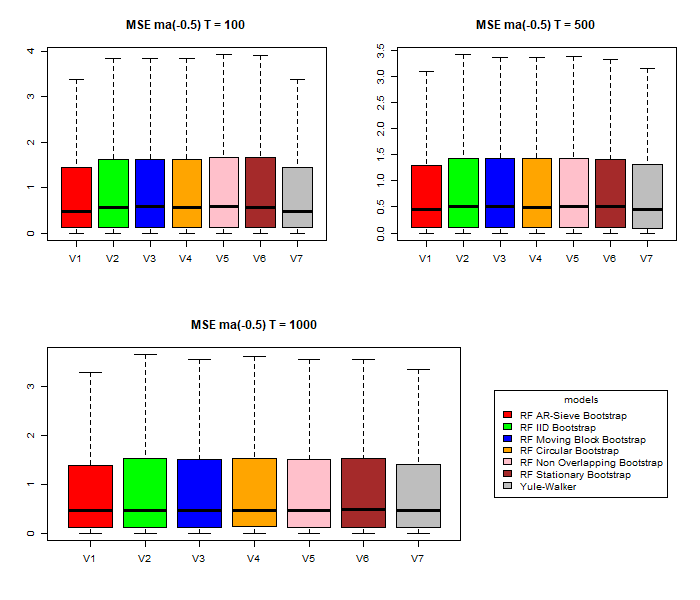}
\caption{\textit{\small MA(1): $\theta_1 = -0.5$}}%
    \label{fig:ma_4_h_1}
\end{figure}

%------------ MA high coefficients

\begin{figure}[H]
    \centering
    \captionsetup{justification=centering}
    
    \includegraphics[width=0.5\textwidth,height=6cm]{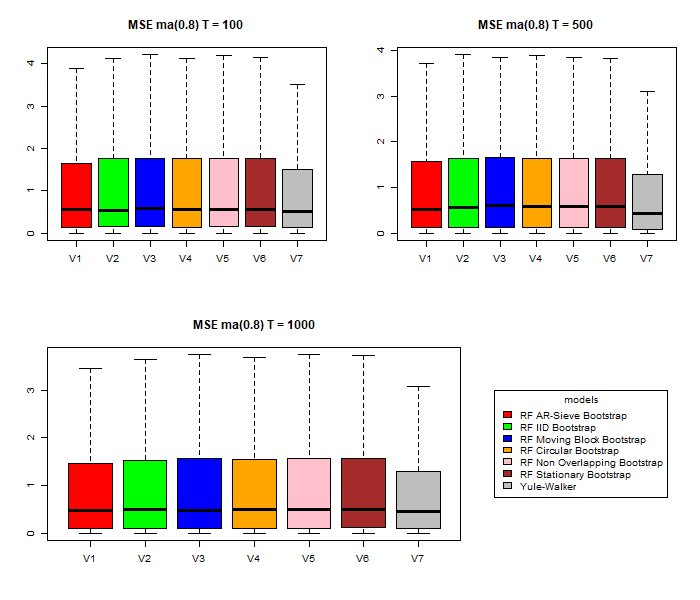}
\caption{\textit{\small MA(1): $\theta_1 = 0.8$}}%
    \label{fig:ma_5_h_1}
\end{figure}

\begin{figure}[H]
    \centering
    \captionsetup{justification=centering}
    
    \includegraphics[width=0.5\textwidth,height=6cm]{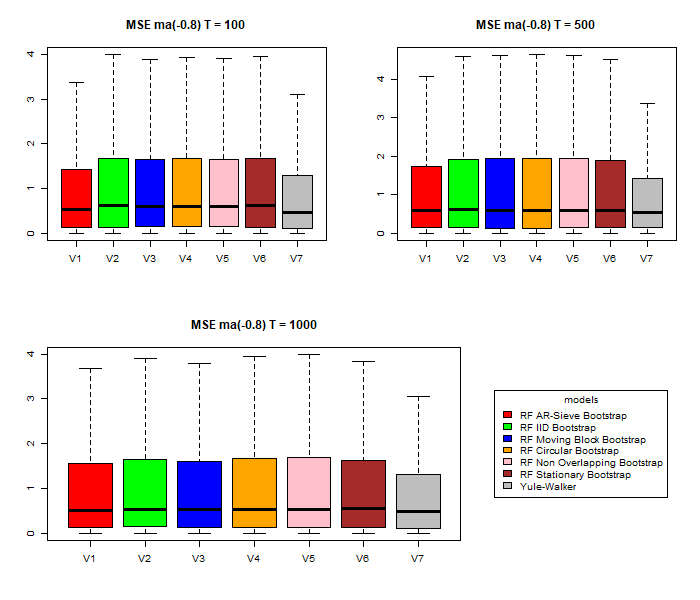}
\caption{\textit{\small AR(1): $\theta_1 = -0.8$}}%
    \label{fig:ma_6_h_1}
\end{figure}

%%%%%%%%%%%%%%%%%%%%%%%%%%%%% ARMA

\subsection{ARMA Models}

\begin{figure}[H]
    \centering
    \captionsetup{justification=centering}
    
    \includegraphics[width=0.5\textwidth,height=6cm]{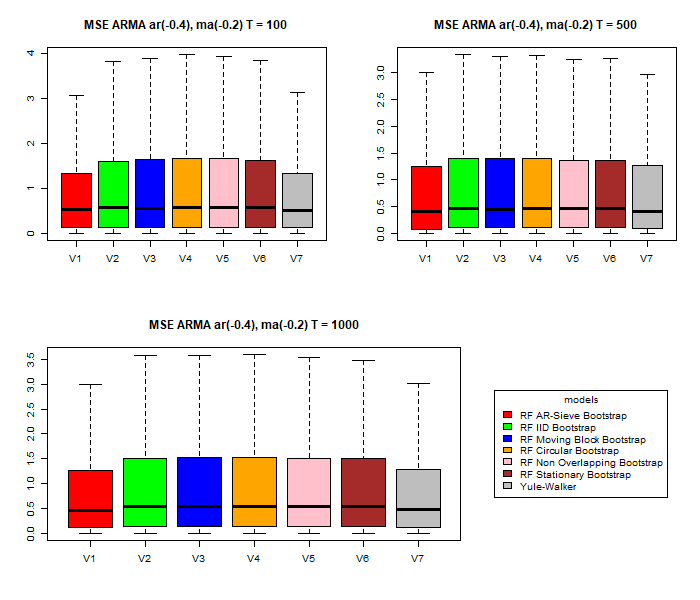}
\caption{\textit{\small ARMA(1): $\phi_1 \!=-0.4,\theta_1=-0.2$}}%
    \label{fig:arma_1_h_1}
\end{figure}

\begin{figure}[H]
    \centering
    \captionsetup{justification=centering}
    
    \includegraphics[width=0.5\textwidth,height=6cm]{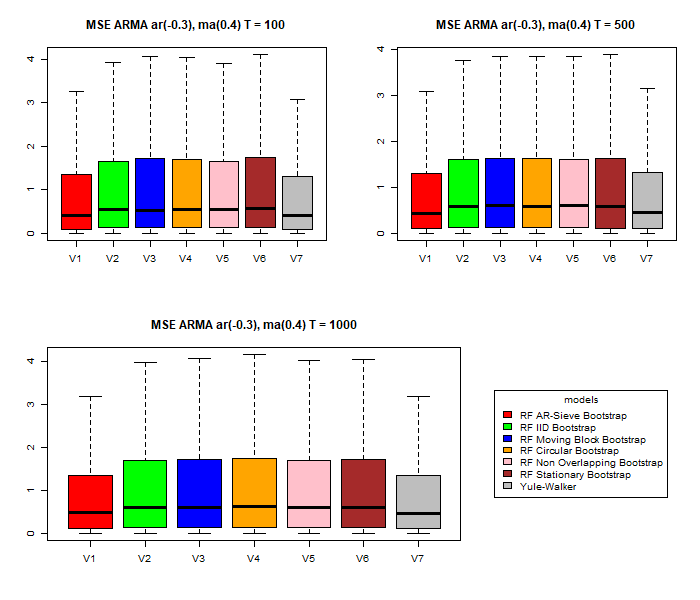}
\caption{\textit{\small ARMA(1,1): $\phi_1=-0.3,\theta_1=0.4$}}%
    \label{fig:arma_2_h_1}
\end{figure}

\begin{figure}[H]
    \centering
    \captionsetup{justification=centering}
    
    \includegraphics[width=0.5\textwidth,height=6cm]{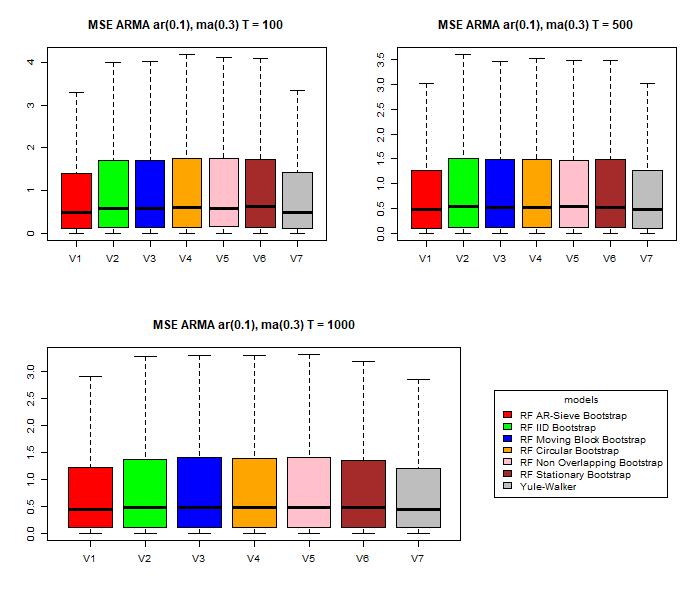}
\caption{\textit{\small ARMA(1,1): $\phi_1=0.1,\theta_1=0.3$}}%
    \label{fig:arma_3_h_1}
\end{figure}

\begin{figure}[H]
    \centering
    \captionsetup{justification=centering}
    
    \includegraphics[width=0.5\textwidth,height=6cm]{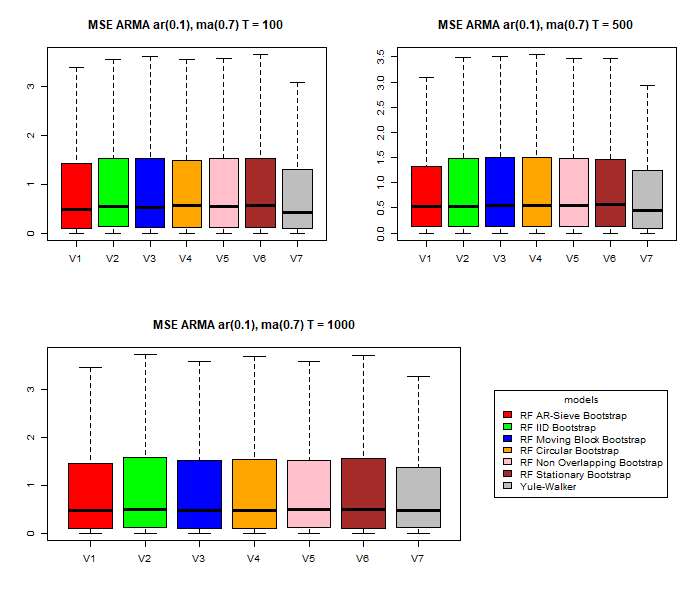}
\caption{\textit{\small ARMA(1,1): $\phi_1=0.1,\theta_1=0.7$}}%
    \label{fig:arma_4_h_1}
\end{figure}

\begin{figure}[H]
    \centering
    \captionsetup{justification=centering}
    
    \includegraphics[width=0.5\textwidth,height=6cm]{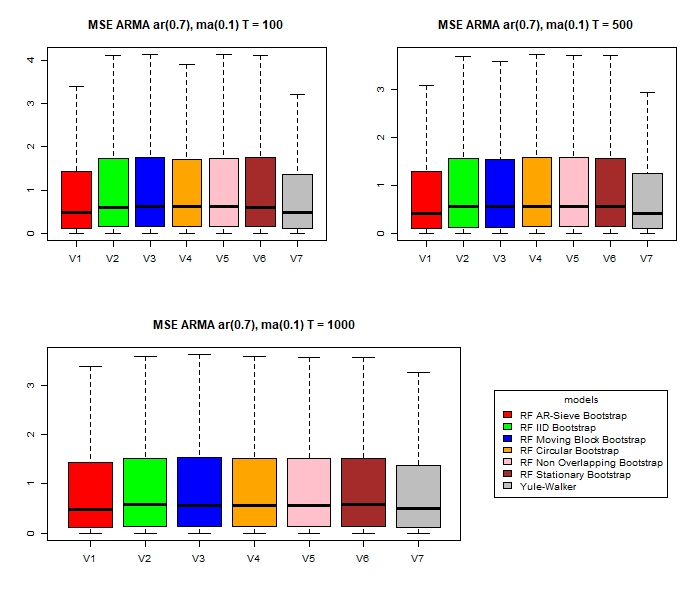}
\caption{\textit{\small ARMA(1,1): $\phi_1=0.7,\theta_1=0.1$}}%
    \label{fig:arma_5_h_1}
\end{figure}

%%%%%%%%%%%%%%%%%%%%%%%%%%%%%ARIMA
\subsection{ARIMA Models}

\begin{figure}[H]
    \centering
    \captionsetup{justification=centering}
    
    \includegraphics[width=0.5\textwidth,height=6cm]{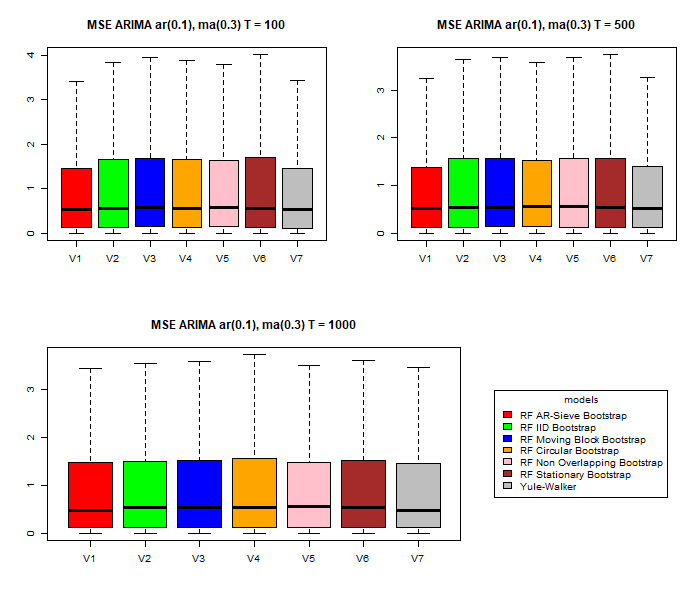}
\caption{\textit{\small ARMA(1,1,1): $\phi_1=0.1,\theta_1=0.3$}}%
    \label{fig:arima_1_h_1}
\end{figure}

\begin{figure}[H]
    \centering
    \captionsetup{justification=centering}
    
    \includegraphics[width=0.5\textwidth,height=6cm]{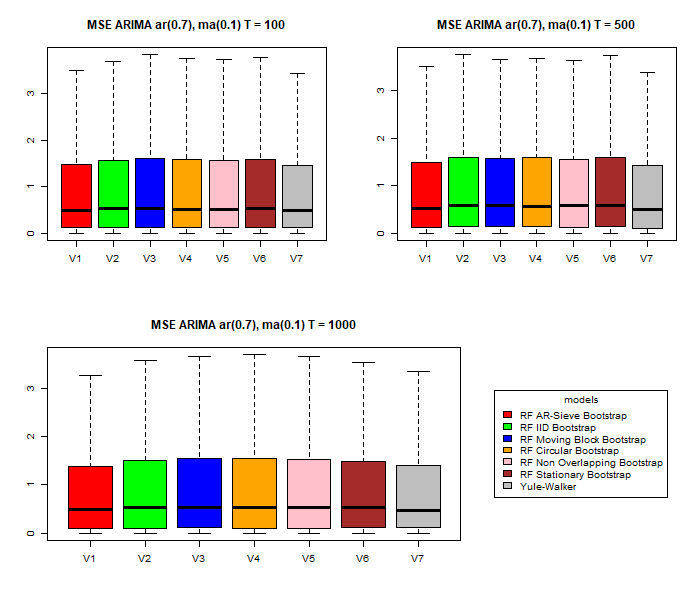}
\caption{\textit{\small ARMA(1,1,1): $\phi_1=0.7,\theta_1=0.1$}}%
    \label{fig:arima_2_h_1}
\end{figure}

\begin{figure}[H]
    \centering
    \captionsetup{justification=centering}
    
    \includegraphics[width=0.5\textwidth,height=6cm]{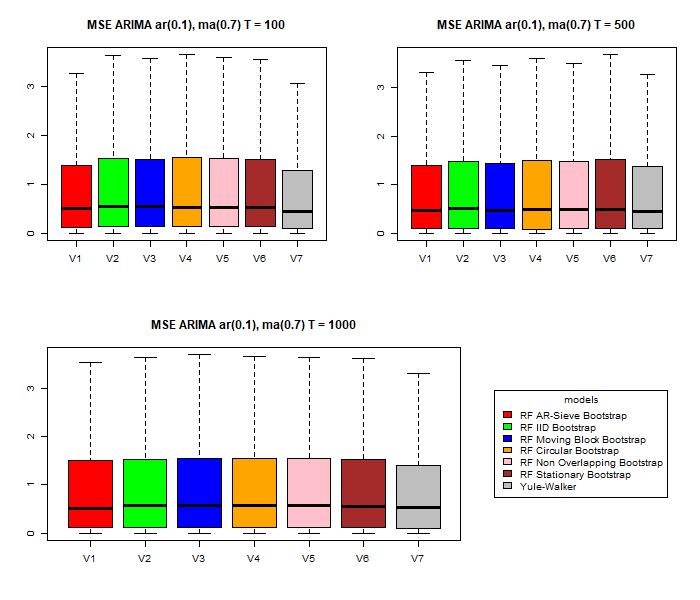}
\caption{\textit{\small ARIMA(1,1,1): $\phi_1=0.1,\theta_1=0.7$}}%
    \label{fig:arima_3_h_1}
\end{figure}

%%%%%%%%%%%%%%%%%%%%%%%%%%%%%ARFIMA
\subsection{ARFIMA Models}

\begin{figure}[H]
    \centering
    \captionsetup{justification=centering}
    
    \includegraphics[width=0.5\textwidth,height=6cm]{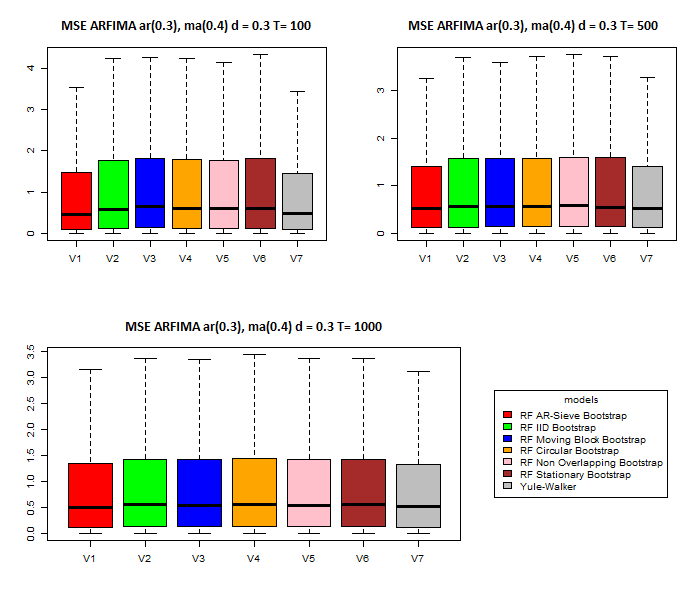}
\caption{\textit{\small ARFIMA(1,1): $\phi_1 = -0.3, \theta_1 = 0.4, d = 0.3$}}%
    \label{fig:arfima_1_h_1}
\end{figure}

\begin{figure}[H]
    \centering
    \captionsetup{justification=centering}
    
    \includegraphics[width=0.5\textwidth,height=6cm]{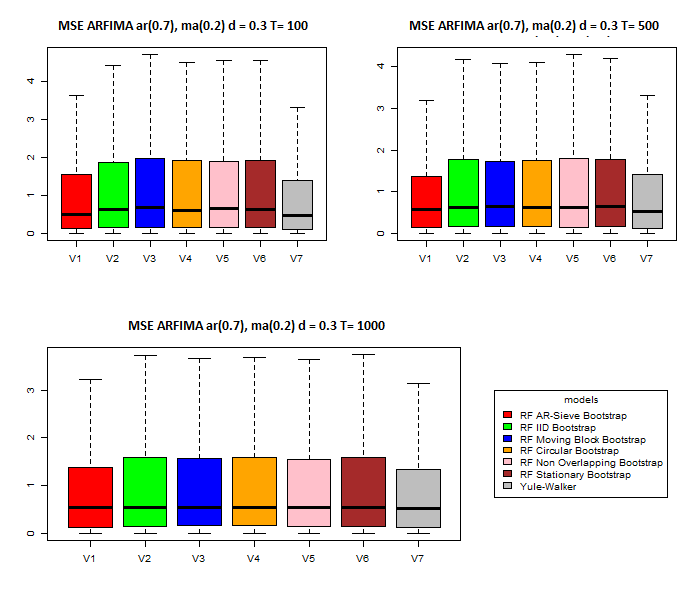}
\caption{\textit{\small ARFIMA(1,1): $\phi_1 = 0.7, \theta_1  = 0.2, d = 0.3$}}%
    \label{fig:arfima_2_h_1}
\end{figure}

%%%%%%%%%%%%%%%%%%%%%%%%%%GARCH

\subsection{GARCH Models}

\begin{figure}[H]
    \centering
    \captionsetup{justification=centering}
    
    \includegraphics[width=0.5\textwidth,height=6cm]{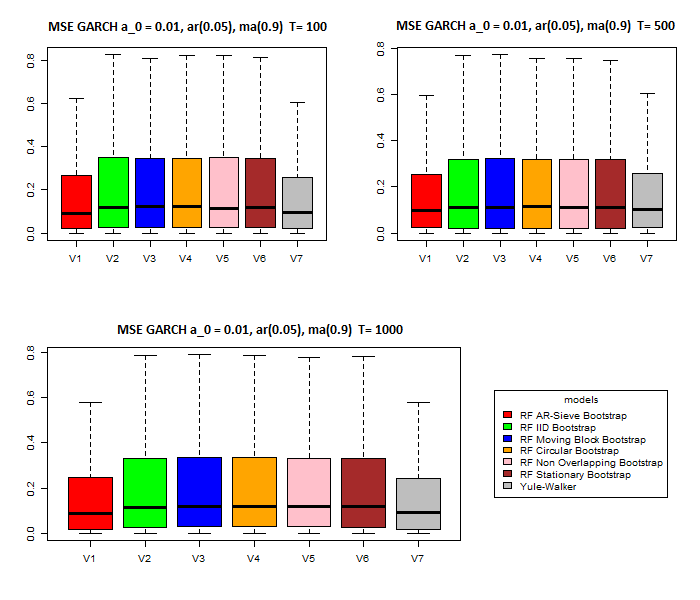}
\caption{\textit{\small GARCH(1,1): $\alpha_0 = 0.01, \alpha_1 = 0.05, \beta_1 = 0.9$}}%
    \label{fig:garch_1_h_1}
\end{figure}

\begin{figure}[H]
    \centering
    \captionsetup{justification=centering}
    
    \includegraphics[width=0.5\textwidth,height=6cm]{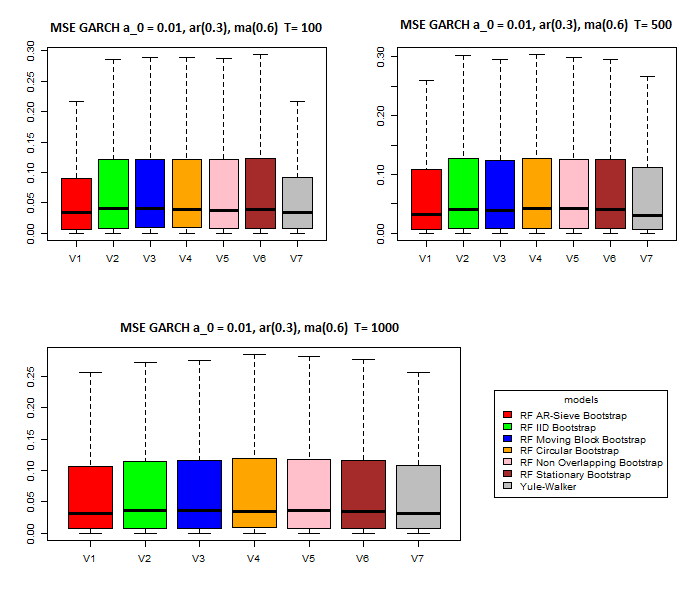}
\caption{\textit{\small GARCH(1,1): $\alpha_0 = 0.01, \alpha_1 = 0.3, \beta_1 = 0.6$}}%
    \label{fig:garch_2_h_1}
\end{figure}

\section{five-step ahead forecast}
\label{app_h_5}
\subsection{AR Models}
\begin{figure}[H]
    \centering
    \captionsetup{justification=centering}
    
    \includegraphics[width=0.5\textwidth, height =6cm]{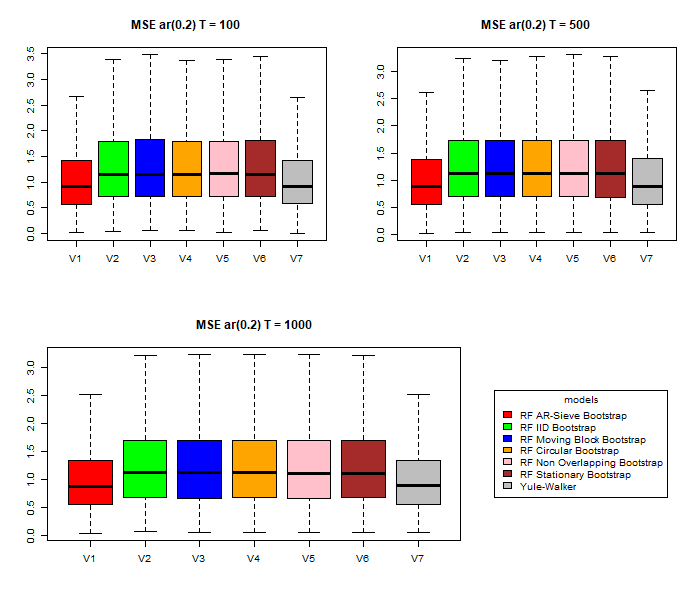}
\caption{\textit{\small AR(1): $\phi_1 = 0.2$}}%
    \label{fig:ar_1_h_5}
\end{figure}

\begin{figure}[H]
    \centering
    \captionsetup{justification=centering}
    
    \includegraphics[width=0.5\textwidth, height = 6cm]{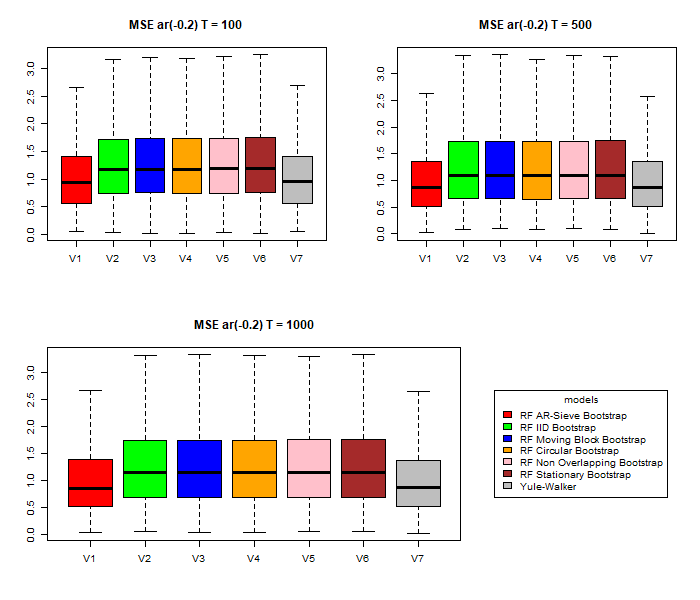}
\caption{\textit{\small AR(1): $\phi_1 = -0.2$}}%
    \label{fig:ar_2_h_5}
\end{figure}

\begin{figure}[H]
    \centering
    \captionsetup{justification=centering}
    
    \includegraphics[width=0.5\textwidth, height=6cm]{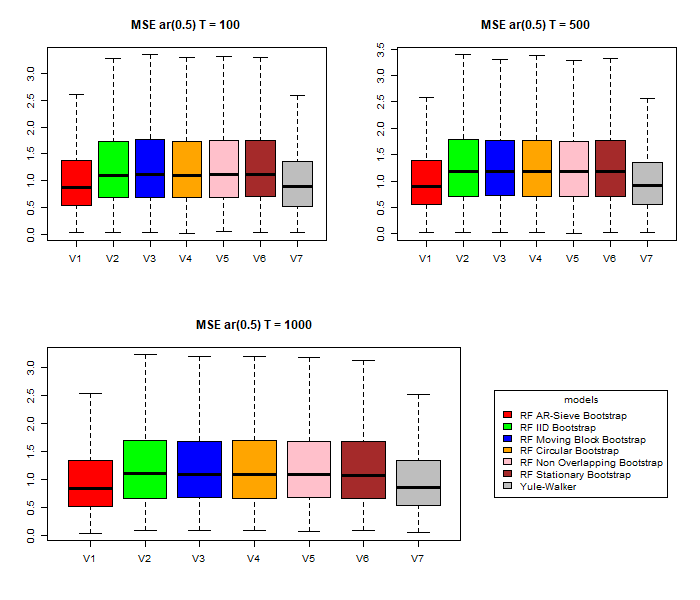}
\caption{\textit{\small AR(1): $\phi_1 = 0.5$}}%
    \label{fig:ar_3_h_5}
\end{figure}

\begin{figure}[H]
    \centering
    \captionsetup{justification=centering}
    
    \includegraphics[width=0.5\textwidth, height=6cm]{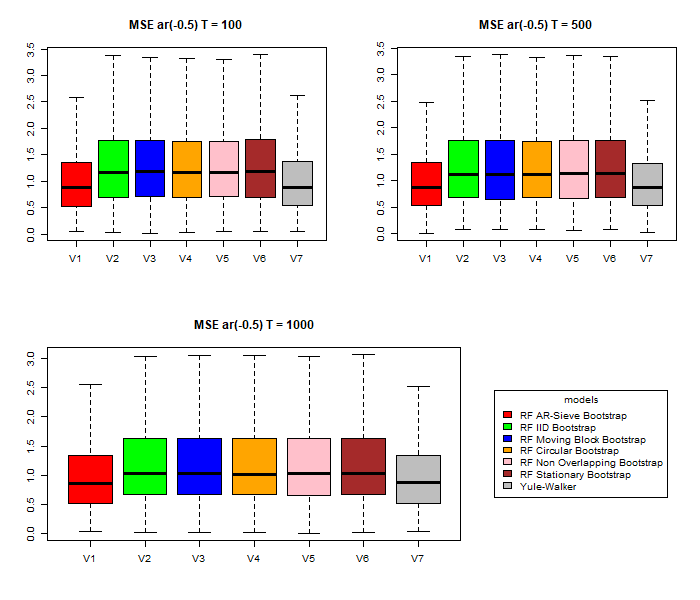}
\caption{\textit{\small AR(1): $\phi_1 = -0.5$}}%
    \label{fig:ar_4_h_5}
\end{figure}

%-------------AR high coefficients

\begin{figure}[H]
    \centering
    \captionsetup{justification=centering}
    
    \includegraphics[width=0.5\textwidth,height=6cm]{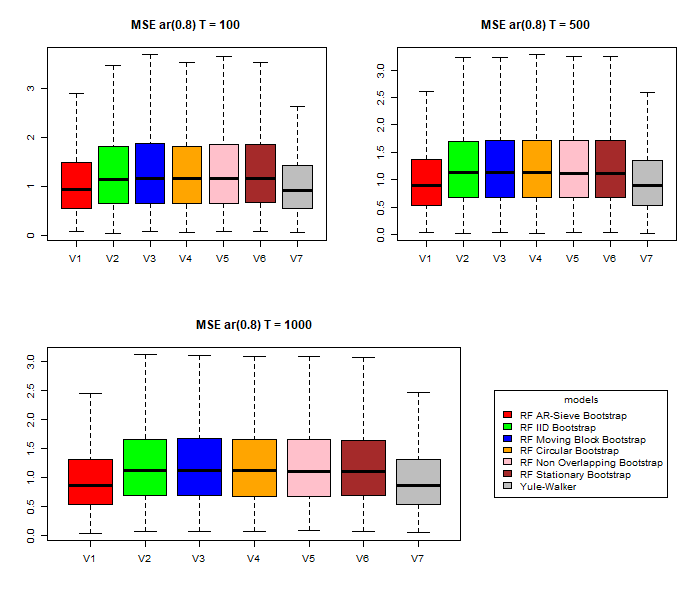}
\caption{\textit{\small AR(1): $\phi_1 = 0.8$}}%
    \label{fig:ar_5_h_5}
\end{figure}

\begin{figure}[H]
    \centering
    \captionsetup{justification=centering}
    
    \includegraphics[width=0.5\textwidth,height=6cm]{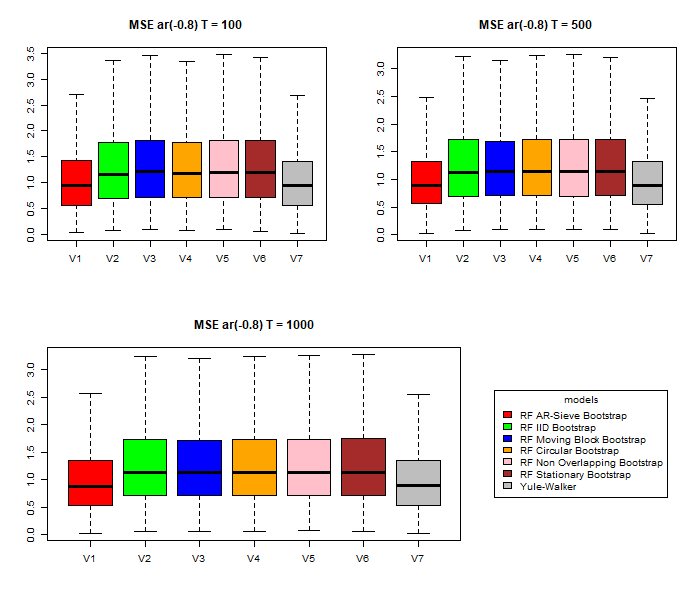}
\caption{\textit{\small AR(1): $\phi_1 = -0.8$}}%
    \label{fig:ar_6_h_5}
\end{figure}

%%%%%%%%%%%%%%%%%%%%%%%%%%%%%%% MA

\subsection{MA Models}
%------------ MA small coefficients

\begin{figure}[H]
    \centering
    \captionsetup{justification=centering}
    
    \includegraphics[width=0.5\textwidth,height=6cm]{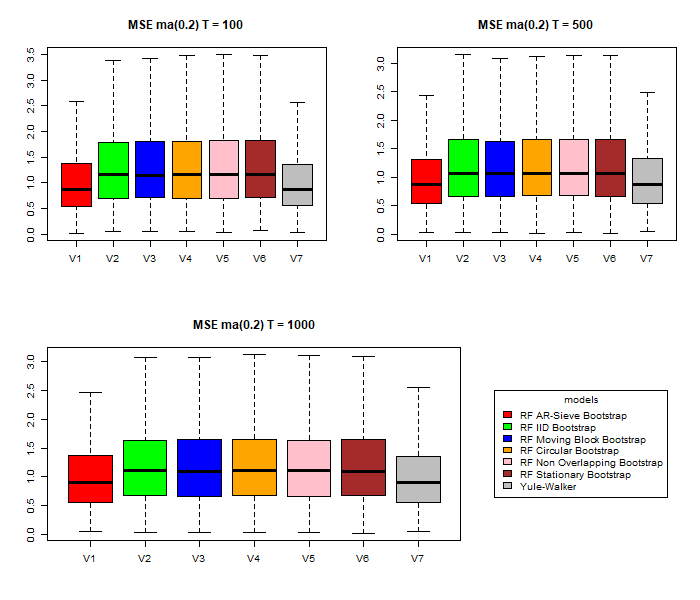}
\caption{\textit{\small MA(1): $\theta_1 = 0.2$}}%
    \label{fig:ma_1_h_5}
\end{figure}

\begin{figure}[H]
    \centering
    \captionsetup{justification=centering}
    
    \includegraphics[width=0.5\textwidth,height=6cm]{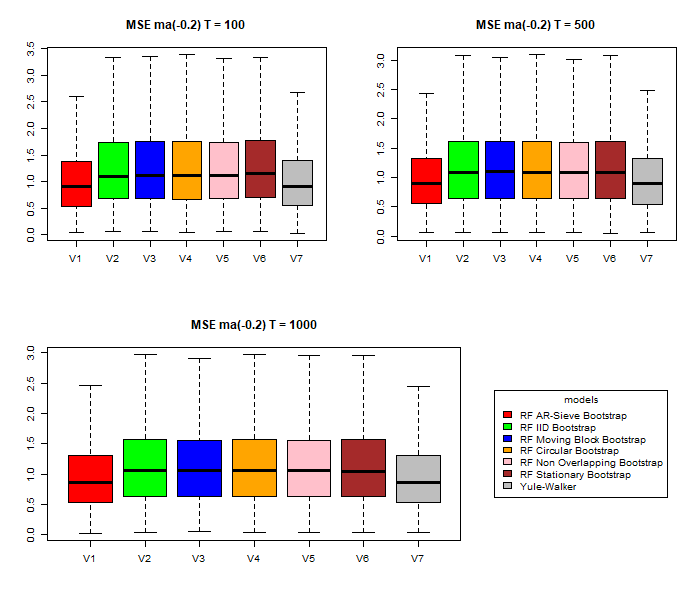}
\caption{\textit{\small MA(1): $\theta_1 = -0.2$}}%
    \label{fig:ma_2_h_5}
\end{figure}

\begin{figure}[H]
    \centering
    \captionsetup{justification=centering}
    
    \includegraphics[width=0.5\textwidth,height=6cm]{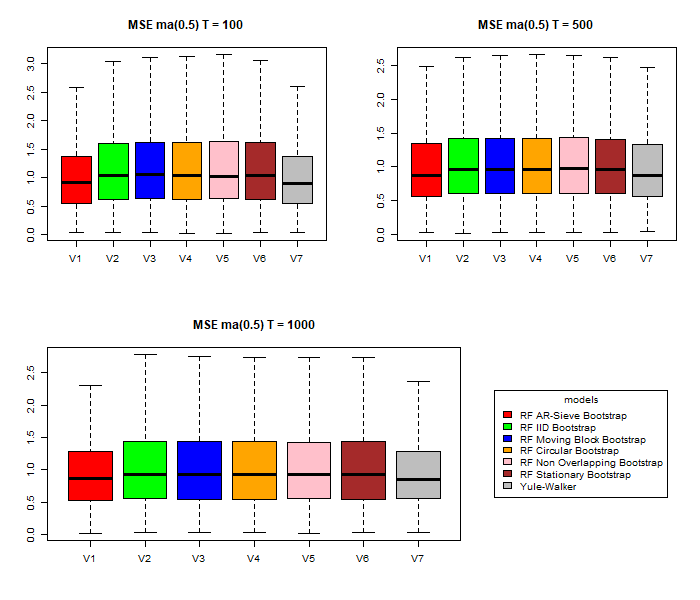}
\caption{\textit{\small MA(1): $\theta_1 = 0.5$}}%
    \label{fig:ma_3_h_5}
\end{figure}

\begin{figure}[H]
    \centering
    \captionsetup{justification=centering}
    
    \includegraphics[width=0.5\textwidth,height=6cm]{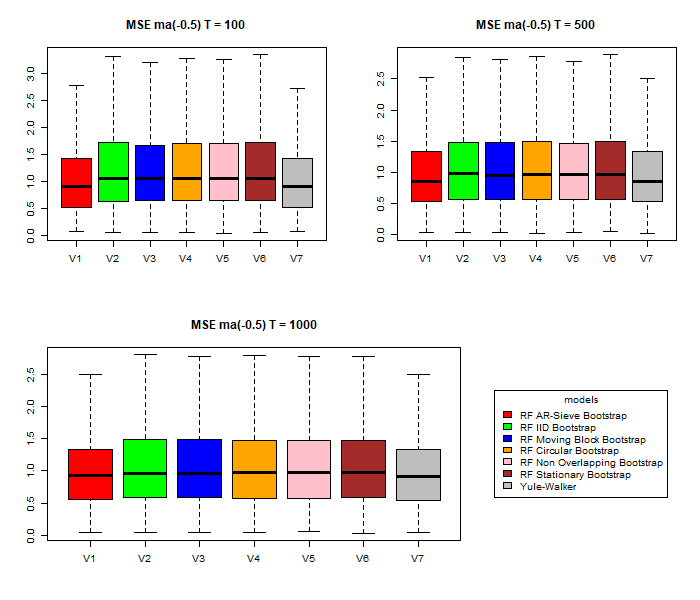}
\caption{\textit{\small MA(1): $\theta_1 = -0.5$}}%
    \label{fig:ma_4_h_5}
\end{figure}

%------------ MA high coefficients

\begin{figure}[H]
    \centering
    \captionsetup{justification=centering}
    
    \includegraphics[width=0.5\textwidth,height=6cm]{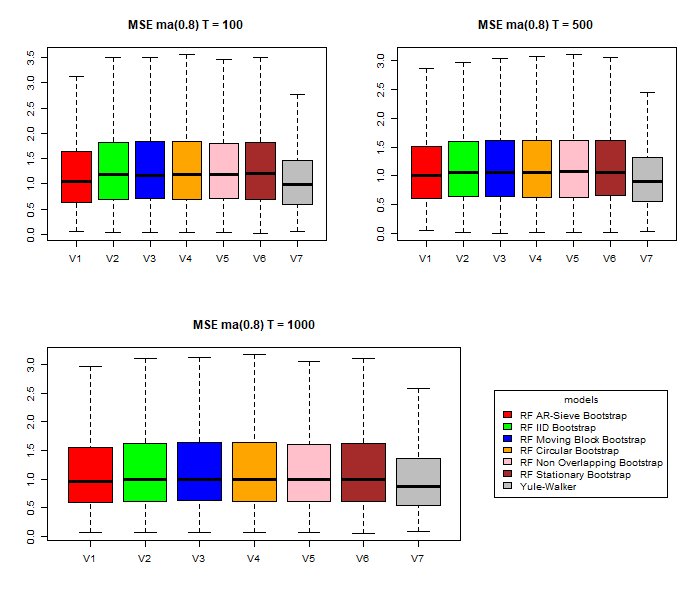}
\caption{\textit{\small MA(1): $\theta_1 = 0.8$}}%
    \label{fig:ma_5_h_5}
\end{figure}

\begin{figure}[H]
    \centering
    \captionsetup{justification=centering}
    
    \includegraphics[width=0.5\textwidth,height=6cm]{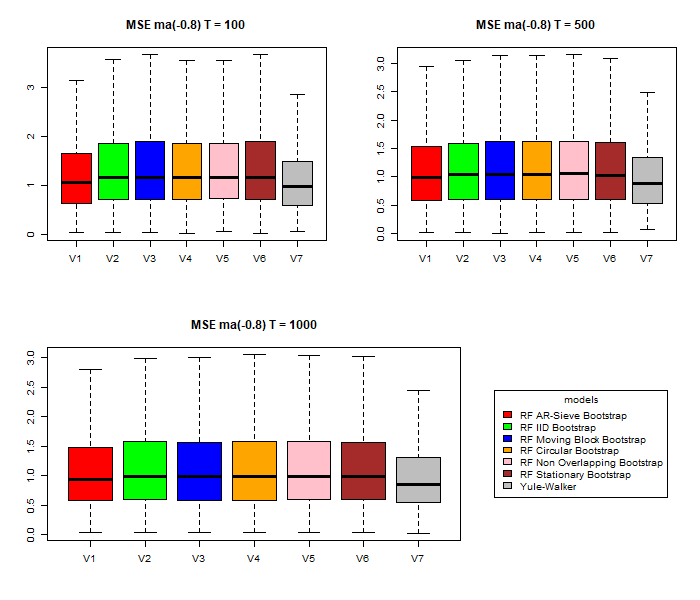}
\caption{\textit{\small AR(1): $\theta_1 = -0.8$}}%
    \label{fig:ma_6_h_5}
\end{figure}

%%%%%%%%%%%%%%%%%%%%%%%%%%%%% ARMA

\subsection{ARMA Models}

\begin{figure}[H]
    \centering
    \captionsetup{justification=centering}
    
    \includegraphics[width=0.5\textwidth,height=6cm]{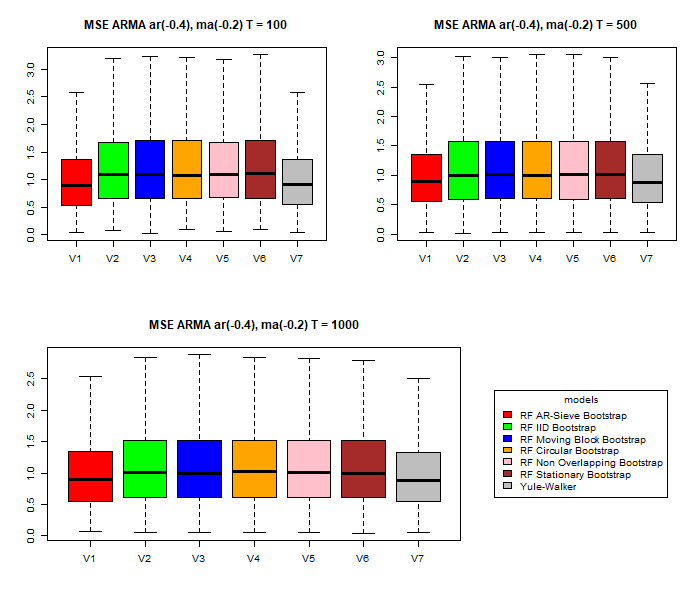}
\caption{\textit{\small ARMA(1): $\phi_1 \!=-0.4,\theta_1=-0.2$}}%
    \label{fig:arma_1_h_5}
\end{figure}

\begin{figure}[H]
    \centering
    \captionsetup{justification=centering}
    
    \includegraphics[width=0.5\textwidth,height=6cm]{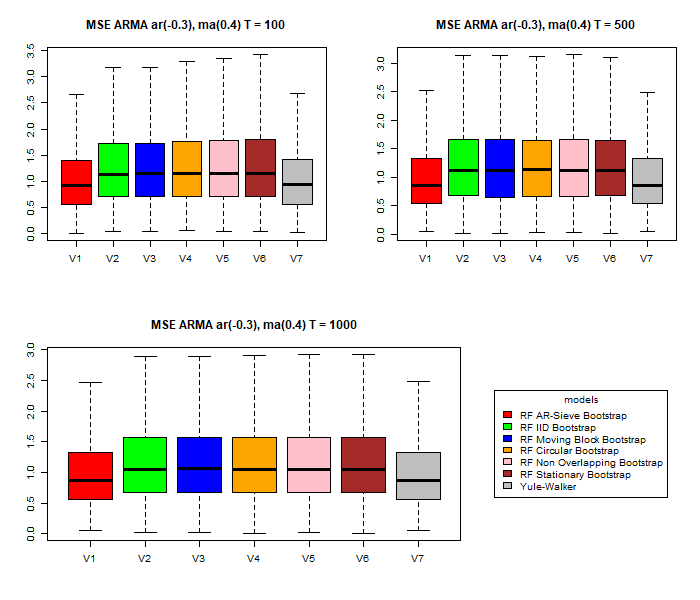}
\caption{\textit{\small ARMA(1,1): $\phi_1=0.3,\theta_1=0.4$}}%
    \label{fig:arma_2_h_5}
\end{figure}

\begin{figure}[H]
    \centering
    \captionsetup{justification=centering}
    
    \includegraphics[width=0.5\textwidth,height=6cm]{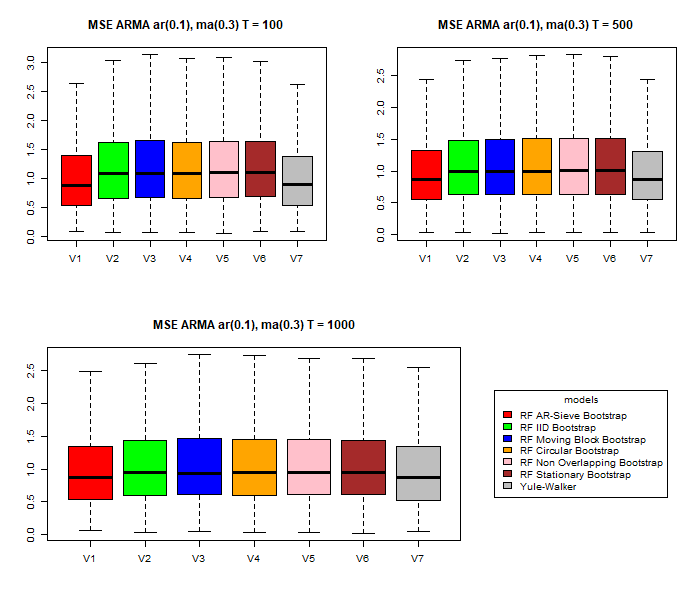}
\caption{\textit{\small ARMA(1,1): $\phi_1=0.1,\theta_1=0.3$}}%
    \label{fig:arma_3_h_5}
\end{figure}

\begin{figure}[H]
    \centering
    \captionsetup{justification=centering}
    
    \includegraphics[width=0.5\textwidth,height=6cm]{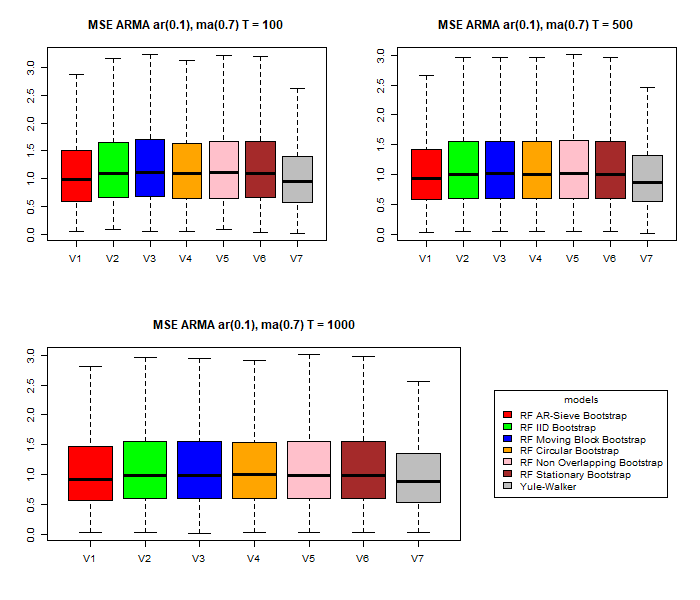}
\caption{\textit{\small ARMA(1,1): $\phi_1=0.1,\theta_1=0.7$}}%
    \label{fig:arma_4_h_5}
\end{figure}

\begin{figure}[H]
    \centering
    \captionsetup{justification=centering}
    
    \includegraphics[width=0.5\textwidth,height=6cm]{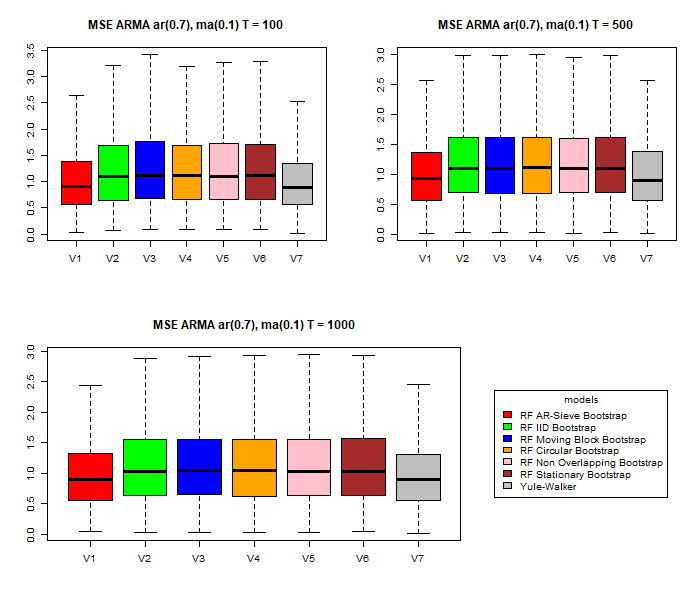}
\caption{\textit{\small ARMA(1,1): $\phi_1=0.7,\theta_1=0.1$}}%
    \label{fig:arma_5_h_5}
\end{figure}

%%%%%%%%%%%%%%%%%%%%%%%%%%%%%ARIMA
\subsection{ARIMA Models}

\begin{figure}[H]
    \centering
    \captionsetup{justification=centering}
    
    \includegraphics[width=0.5\textwidth,height=6cm]{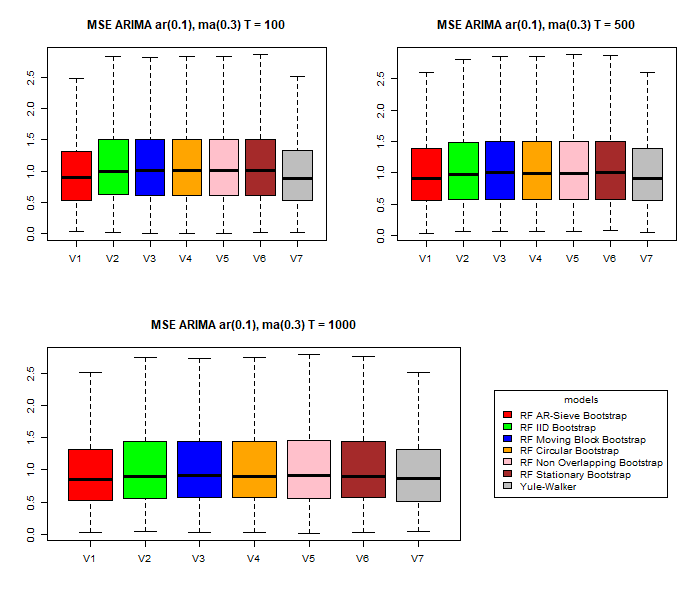}
\caption{\textit{\small ARMA(1,1,1): $\phi_1=0.1,\theta_1=0.3$}}%
    \label{fig:arima_1_h_5}
\end{figure}

\begin{figure}[H]
    \centering
    \captionsetup{justification=centering}
    
    \includegraphics[width=0.5\textwidth,height=6cm]{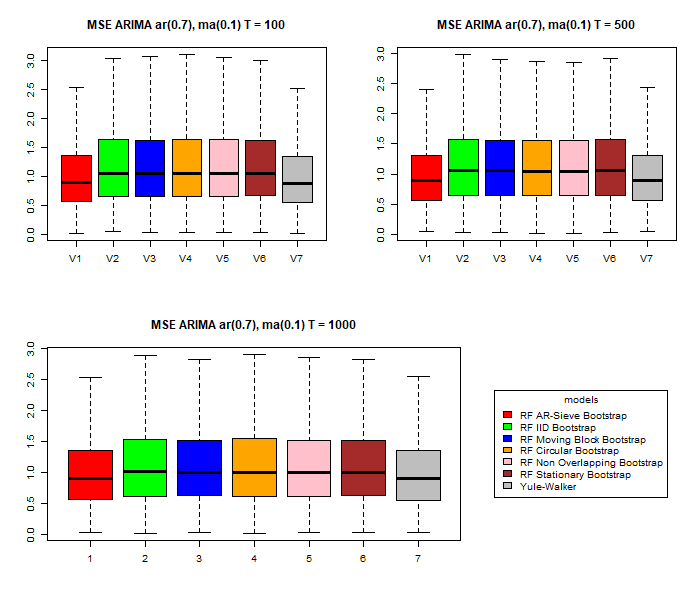}
\caption{\textit{\small ARMA(1,1,1): $\phi_1=0.7,\theta_1=0.1$}}%
    \label{fig:arima_2_h_5}
\end{figure}

\begin{figure}[H]
    \centering
    \captionsetup{justification=centering}
    
    \includegraphics[width=0.5\textwidth,height=6cm]{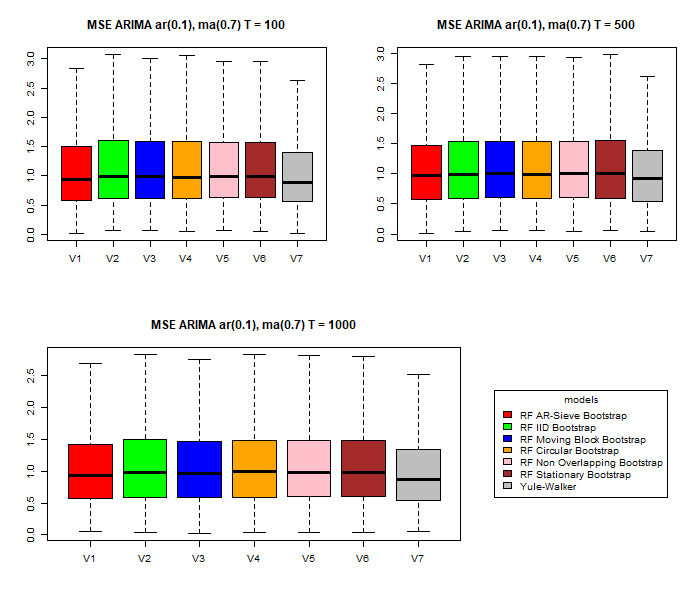}
\caption{\textit{\small ARIMA(1,1,1): $\phi_1=0.1,\theta_1=0.7$}}%
    \label{fig:arima_3_h_5}
\end{figure}

%%%%%%%%%%%%%%%%%%%%%%%%%%%%%ARFIMA
\subsection{ARFIMA Models}

\begin{figure}[H]
    \centering
    \captionsetup{justification=centering}
    
    \includegraphics[width=0.5\textwidth,height=6cm]{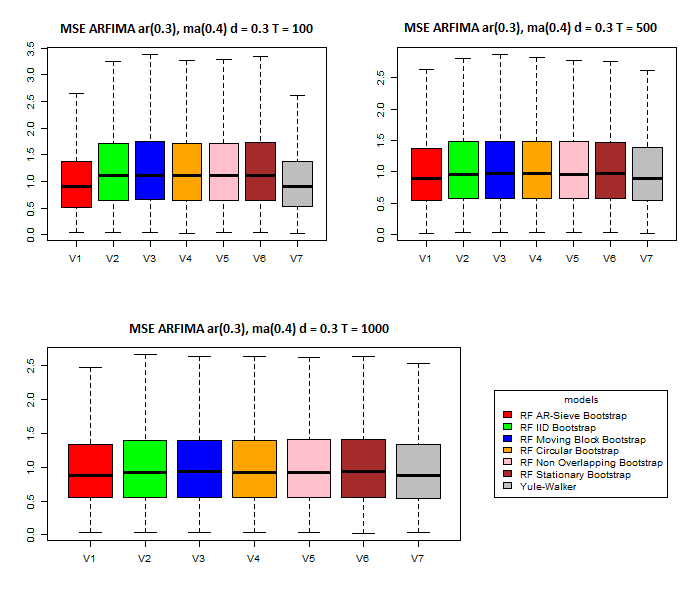}
\caption{\textit{\small ARFIMA(1,1): $\phi_1 = -0.3, \theta_1 = 0.4, d = 0.3$}}%
    \label{fig:arfima_1_h_5}
\end{figure}

\begin{figure}[H]
    \centering
    \captionsetup{justification=centering}
    
    \includegraphics[width=0.5\textwidth,height=6cm]{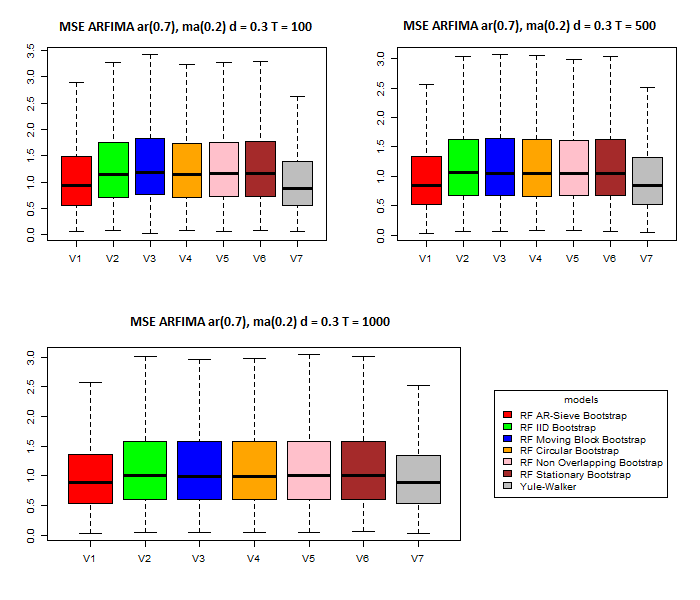}
\caption{\textit{\small ARFIMA(1,1): $\phi_1 = 0.7, \theta_1  = 0.2, d = 0.3$}}%
    \label{fig:arfima_2_h_5}
\end{figure}

%%%%%%%%%%%%%%%%%%%%%%%%%%GARCH

\subsection{GARCH Models}

\begin{figure}[H]
    \centering
    \captionsetup{justification=centering}
    
    \includegraphics[width=0.5\textwidth,height=6cm]{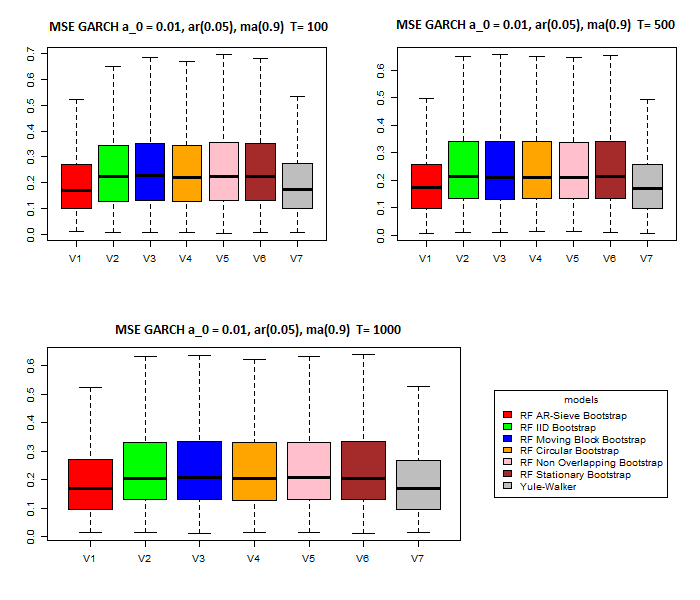}
\caption{\textit{\small GARCH(1,1): $\alpha_0 = 0.01, \alpha_1 = 0.05, \beta_1 = 0.9$}}%
    \label{fig:garch_1_h_5}
\end{figure}

\begin{figure}[H]
    \centering
    \captionsetup{justification=centering}
    
    \includegraphics[width=0.5\textwidth,height=6cm]{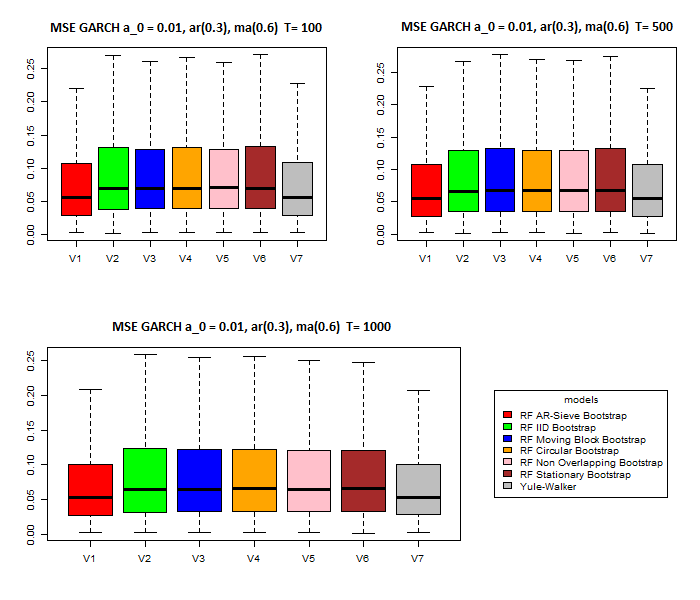}
\caption{\textit{\small GARCH(1,1): $\alpha_0 = 0.01, \alpha_1 = 0.3, \beta_1 = 0.6$}}%
    \label{fig:garch_2_h_5}
\end{figure}

\end{document}